\documentclass[twoside,11pt]{article}

\usepackage[abbrvbib, preprint]{dmlrarxiv}

\usepackage[T1]{fontenc}

\usepackage{amsmath}
\usepackage{subcaption}
\usepackage[most]{tcolorbox}

\usepackage{fvextra}
\DefineVerbatimEnvironment{BreakVerbatim}{Verbatim}{
    breaklines=true,
    breakanywhere=true,
    breaksymbol={},
    breaksymbolindent=0pt,
    commandchars=\\\{\},  
}

\usepackage{booktabs}  
\usepackage[table]{xcolor}

\usepackage{xcolor}
\usepackage{listings}  

\definecolor{codegreen}{rgb}{0,0.6,0}
\definecolor{codegray}{rgb}{0.5,0.5,0.5}
\definecolor{codepurple}{rgb}{0.58,0,0.82}
\definecolor{backcolour}{rgb}{0.95,0.95,0.92}

\lstdefinestyle{mystyle}{
    basicstyle=\fontsize{8}{8}\selectfont\ttfamily,
    commentstyle=\color{codegreen}, 
    keywordstyle=\color{black},
    numberstyle=\tiny\color{codegray},
    stringstyle=\color{codepurple},
    breakatwhitespace=false,         
    breaklines=true,                 
    captionpos=b,                    
    keepspaces=true,                 
    numbersep=0pt,                  
    showspaces=false,                
    showstringspaces=false,
    showtabs=false,                  
    tabsize=2
}
 
\lstdefinestyle{pythoncode}{
  language=Python,
  basicstyle=\ttfamily\small,
  backgroundcolor=\color{gray!5},
  frame=single
}

\lstdefinestyle{output}{
  basicstyle=\ttfamily\scriptsize,
  backgroundcolor=\color{gray!10},
  frame=single,
  showstringspaces=false
}

\usepackage{lastpage}
\dmlrheading{23}{2025}{1-\pageref{LastPage}}{9/25; Revised 10/25}{11/25}{21-0000}{Hui Wen Goh and Jonas Mueller} 

\ShortHeadings{Trustworthiness Scoring for LLM Structured Outputs and Data Extraction}{Goh and Mueller}
\firstpageno{1}

\begin{document}

\title{Real-Time Trustworthiness Scoring for LLM Structured Outputs and Data Extraction}

\author{\name Hui Wen Goh \email huiwen@cleanlab.ai \\
       \addr Cleanlab\\
       \AND
       \name Jonas Mueller \email jonas@cleanlab.ai \\
       \addr Cleanlab
}

\editor{My editor}

\maketitle

\vspace*{-11.5mm}
\begin{abstract}
Structured Outputs from current LLMs exhibit sporadic errors, hindering enterprise AI deployment. We present CONSTRUCT, a real-time uncertainty estimator that scores the trustworthiness of LLM Structured Outputs. Lower-scoring outputs are more likely to contain errors, enabling automatic prioritization of limited human review bandwidth. CONSTRUCT additionally scores the trustworthiness of each field within a Structured Output, helping reviewers quickly identify which parts of the output are incorrect. Our method is suitable for any LLM (including black-box LLM APIs without logprobs), does not require labeled training data or custom model deployment, and supports complex Structured Outputs with heterogeneous fields and nested JSON schemas.

We also introduce one of the first public LLM Structured Output benchmarks with reliable ground-truth values. Over this four-dataset benchmark, CONSTRUCT detects errors in outputs from various LLMs (including Gemini 3 and GPT-5) with significantly higher precision/recall than existing techniques.
\end{abstract}

\section{Introduction}

Significant enterprise value from LLMs comes from their ability to produce
\emph{Structured Outputs}, turning raw/unstructured information into structured data suitable for business use.  
In \emph{Structured Outputs} mode, LLM APIs ensure that their outputs are organized into specific named \emph{fields} and adhere to a provided schema that dictates a specific type for the \emph{value} in each field \citep{openai2024structuredoutputs}.  
This capability can even produce complex outputs with \emph{nested} JSON structures.  
  Applications of LLM Structured Outputs include: data extraction, document processing, and seamless handoff between adaptive LLM intelligence and deterministic software logic \citep{balasubramanian2025leveraging, schick2023toolformer}.  

We introduce \textbf{CONSTRUCT} (CONfidence scoring for STRUCTured Outputs) -- a method for scoring how much one can \emph{trust} any Structured Output generated from any LLM, as well as each field within that output. \emph{Trustworthiness scores} range between 0 to 1, such that \textbf{per-field} trustworthiness scores near 1 indicate fields whose value is likely to be correct and \textbf{per-document} trustworthiness scores near 1 indicate outputs whose fields are likely all correct (supposing the LLM produces one structured output per input document).
CONSTRUCT is being used by finance and insurance companies, and was motivated by the following repeated observations made while working with them:

Since the launch of ChatGPT, many teams manually handling labor-intensive document processing workflows have investigated LLM automation. While initial LLM prototypes seem promising thanks to the Structured Outputs capability, teams often find that at scale, they are unable to rely on LLM automation due to sporadic errors in these LLM Structured Outputs and their inability to handle edge-cases/outliers. 
Effective \emph{trustworthiness scoring} of each LLM output enables these teams to prioritize which documents warrant human review (as well as which specific fields in the structured output to review). This enables \emph{human-in-the-loop workflows} in which teams manually handle 1-5\% of the cases where LLMs are untrustworthy, while the remaining 95-99\% of their work is reliably LLM-automated.

Below is a concrete example of using CONSTRUCT via our open-source Python code\footnote{Code to run CONSTRUCT trustworthiness scoring:  \url{https://github.com/cleanlab/tlm}}. Note how our method  helps automatically: catch erroneous Structured Outputs, discern which specific fields are wrong, and explain why they are untrustworthy.

\begin{quote}
\begin{lstlisting}[language=Python] 
prompt = 'Extract the vendor name, invoice date, total amount, and currency from the following invoice:  Payment of $1,530.00 USD was issued to Brightstone Manufacturing for the invoice dated February 12, 2024.'

openai_kwargs = {
    'messages': [{'role': 'user', 'content': prompt}],
    'response_format': MyOutputSchema,  # Pydantic/JSON schema
    ...
}

llm_output = OpenAI.chat.completions.parse(**openai_kwargs)  # generate

llm_output['content']
\end{lstlisting}
\vspace*{-0.5em}
\begin{lstlisting}[style=output]
{
  'vendor': 'Brightstone Manufacturing',
  'invoice_date': '2024-02-31',
  'total_amount': '1530.50',
  'currency': 'USD'
}
\end{lstlisting}

\begin{lstlisting}[language=Python] 
result = CONSTRUCT.score(llm_output, **openai_kwargs)

result['trustworthiness_score']
\end{lstlisting}
\vspace*{-0.5em}
\begin{lstlisting}[style=output]
  0.18
\end{lstlisting}

\begin{lstlisting}[language=Python] 
result['untrustworthy_fields']
\end{lstlisting}
\vspace*{-0.5em}
\begin{lstlisting}[style=output]
Untrustworthy fields: ['invoice_date', 'total_amount']

Field: invoice_date
Response: 2024-02-31
Score: 0.03
Explanation: The extracted date '2024-02-31' is invalid and does not match the actual date 'February 12, 2024' in the text.

Field: total_amount
Response: 1530.50
Score: 0.16
Explanation: The extracted amount ('1530.50') does not match the correct total ('$1,530.00').
\end{lstlisting}

\begin{lstlisting}[language=Python] 
result['per_field_scores'] 
\end{lstlisting}
\vspace*{-0.5em}
\begin{lstlisting}[style=output]
{
  'vendor': 0.99,
  'invoice_date': 0.03,
  'total_amount': 0.16,
  'currency': 0.98
}
\end{lstlisting}
\end{quote}


Without overall (per-document) trustworthiness scores, one might easily assume the LLM output is accurate, even for this example that is much simpler than enterprise document processing tasks.  Without per-field scores and automated explanations, it would take much more time to review documents receiving low overall trustworthiness scores and figure out what the errors are (especially when there are 20+ fields per output).

To address these needs, our paper makes three main contributions:
\begin{enumerate}
\item A real-time method to score the trustworthiness of Structured Outputs from \emph{any} LLM, with strong precision/recall for detecting erroneous outputs.
\item A new Structured Outputs benchmark\footnote{Our Structured Outputs Benchmark:  \url{https://github.com/cleanlab/structured-output-benchmark}} composed of 4 datasets that we intensively curated.  Unlike existing public benchmarks in this area, these datasets are not full of mistakes/ambiguities preventing reliable LLM evaluation.
\item Comprehensive experiments that benchmark the performance of 5 frontier LLMs and 4 trust scoring techniques applied to detect errors from each model.
\end{enumerate}

\section{Related Work}
\label{sec:relatedwork}

One approach to detect LLM output errors would be to train a model for this task \citep{kim2024prometheus, ravi2024lynx}.  This requires intensive data preparation/labeling work, LLM-scale infrastructure for custom model serving, and detector models need to be retrained whenever new LLMs are released (which make different/harder-to-detect types of errors).
The enterprises that we surveyed did not want to deal with this, as teams were occupied getting their  LLM to generate good outputs. 
Here we instead focus on what these teams preferred: trustworthiness scoring methods that work \textbf{out-of-the-box}, which do not require additional compute infrastructure beyond the same LLM API used to generate the Structured Outputs. Such methods can utilize \textbf{any} pretrained, off-the-shelf LLM as a \emph{verifier} model instead of custom models (e.g.\ frontier LLMs from OpenAI, Anthropic, Gemini, etc).  
The scores from such methods instantly improve whenever LLMs advance, simply by upgrading this verifier LLM \citep{sutton2019bitter}.

\emph{LLM-as-a-Judge} is one such approach to assess LLM outputs in reference-free evaluations \citep{li2024llmjudges}.  Here, the same LLM (or a different model) is prompted to directly evaluate the original LLM's response. Standard approaches are to ask the model to ``double-check'' whether the response is \emph{correct} or to rate its \emph{confidence} in this response \citep{zheng2023judgingllmasajudgemtbenchchatbot}.  While such self-assessment may conceptually seem questionable, it works well enough to be widely used across enterprise LLM applications \citep{kadavath2022language,yang2024verbalized,lee2024rlaif}. 
Since the original realization by \citet{tian2023just} that \emph{verbalized confidence} can detect LLM errors with surprising efficacy, models as verifiers have consistently improved and we expect the trend to continue as verification tasks are increasingly incorporated into the Frontier models’ training data.

Since LLMs are autoregressive probabilistic models even when generating Structured Outputs, another popular confidence-scoring approach relies on the model's estimated probability of its generated response. The specific measure typically used is the \emph{average log probability of the tokens} generated by the model  (\emph{logprobs}).\footnote{Since many regard logprobs as   \emph{the confidence score} for machine learning, we call our method's output a \emph{trustworthiness score} \citep{pmlr-v235-huang24x}. Logprobs has issues in open-domain LLM applications that did not exist in classical machine learning and is not a complete solution to LLMs' major trust problem.  Confidence scores are also associated with \emph{calibration} while we focus on \emph{error-detection} \citep{ni2019calibration}.
}
However, these \emph{logprobs} can fail to account for \emph{epistemic uncertainty} as well as the fact that there are many ways to express the same semantics as a sequence of tokens \citep{farquhar2024semanticentropy}.  Furthermore, many LLM APIs do not expose the model's logprobs, such as Anthropic models, or certain reasoning models from OpenAI/Google \citep{srivastava2025cracking, sophiabits2024leveraging}.

Various other methods have been proposed for scoring the trustworthiness of LLM outputs, often for  \emph{hallucination detection}, but to our knowledge, these methods have not been designed/used for Structured Outputs \citep{farquhar2024semanticentropy, liu2025uncertainty}.  Like the BSDetector method of \citet{chen2024bsdetector}, CONSTRUCT also aggregates various forms of LLM-derived evidence into an overall trustworthiness score.

 \section{Scoring the Trustworthiness of Structured Outputs from any LLM}
\label{sec:method}


We consider Structured Outputs applications where each example involves a \emph{document} from which the LLM should extract certain data \citep{docetl}.  
Let $x$ denote an input \emph{prompt}, composed of instructions about the task at hand plus the document to process. Let $\hat{y}$ denote the corresponding \emph{structured output} generated by the LLM, which adheres to a prespecified \emph{schema}.  Specified via JSON/Pydantic as another input to the LLM, the schema defines a set of $n$ top-level \emph{fields} that must be present in the output:
$
\hat{y} = [ \hat{y}_1, \hat{y}_2, \dots, \hat{y}_n ].
$
CONSTRUCT works for arbitrary structured outputs whose fields $ \hat{y}_i$ may include: numbers, categories, lists, text, and further nested JSON with sub-fields of  complex types.

CONSTRUCT can employ any pretrained, off-the-shelf LLM (call this the \emph{verifier} model) to score the trustworthiness of:  structured outputs from any LLM (call this the \emph{generator} model) as well as each field within a structured output.
In document-processing applications, the documents with the lowest \textbf{per-document} trust scores are likely to contain some error in their corresponding LLM structured output, while the \textbf{per-field} trust scores reveal exactly where likely errors occur.   CONSTRUCT only takes in the same input given to the generator LLM (prompt + schema) as well as its generated output.
The challenge is to detect erroneous outputs with \textbf{high precision/recall} and \textbf{low latency}, which limits the total number/sequence of verifier LLM calls that can be made.  We designed CONSTRUCT to be \textbf{general-purpose} and \textbf{simple}, leading to critical advantages listed in Section \ref{sec:discussion} that make this method suitable for the enterprises we surveyed. 

\subsection{Principles for an effective trustworthiness score}

A straightforward way to use the verifier LLM for \textbf{per-document} scoring is via \textbf{LLM-as-a-Judge} to comprehensively evaluate $(x, \hat{y})$.  
Ways one could use the verifier LLM for \textbf{per-field} scoring include two LLM-as-a-Judge variants:  
\begin{enumerate}
\item  \textbf{Multiple Calls.} Use a separate LLM-Judge to score each individual field independently of the others (one score per LLM call).
\item \textbf{Single Call.} Run one LLM-Judge call yielding $n$ scores, where the verifier LLM simultaneously scores all fields by producing a JSON Structured Output of scores.
\end{enumerate}
\emph{Multiple Calls} is slower/costlier than the \emph{Single Call} LLM-as-a-Judge, and one enterprise told us they encountered token-rate-limit API problems when attempting to use this approach in financial document processing applications involving 50+ output fields.  Thus, CONSTRUCT avoids this non-scalable \emph{Multiple Calls} technique, yet still outperforms it. 



 
 CONSTRUCT builds on the LLM-as-a-Judge approach to per-document scoring and Single-Call per-field scoring, in a way that overcomes the following \textbf{shortcomings of capacity-constrained approximate information processing} in transformer-based structured output generation/verification: 
\begin{itemize}
\item[S1)] Single-Call per-field LLM-as-a-Judge is prone to misallocating information-processing capacity across output fields.
\item[S2)] Per-document LLM-as-a-Judge is prone to overlooking mistakes in some fields. 
\item[S3)] Prescribing how individual LLM calls should perform their assessment violates the \emph{bitter lesson} of \citet{sutton2019bitter}, but omitting tips or intermediate-steps drops performance.
\item[S4)] The quality of outputs from LLM calls degrades if we ask for too much information.
\item[S5)] One LLM call (neural-network-based  autoregressive generation of a single response) is unreliable and can yield unpredictable mistakes.
\end{itemize}
To understand S1 and S2, consider verifying one particular field in a structured output (as in the \emph{Multiple Calls} approach).  Here the verifier  model's self-attention layers must allocate the right information processing capacity to: (i) the part of the output being scored containing the predicted value for this field ($\hat{y}_i$), (ii) the part of the schema governing this field, (iii) the part of the prompt containing relevant instructions for getting this field right, (iv) the part of the document with the relevant information.  If we instead ask the verifier LLM to output verifications of all fields within one forward pass through the neural network (as in the \emph{Single Call} approach), its self-attention layers must further divide the aforementioned information processing capacity appropriately between the fields.
Since a pretrained, off-the-shelf LLM will not have learned which fields/values to allocate more processing capacity for, the \emph{Single Call} LLM-as-a-Judge empirically performs much worse than  \emph{Multiple Calls} (Figure \ref{fig:auroc_field}).
Alternatively, one can let the verifier LLM adaptively decide how to allocate processing capacity across fields, as in the holistic Per-Document LLM-as-a-Judge above. However, we found that this freedom causes the verifier to systematically overlook mistakes in certain fields, which are caught by the \emph{Single Call} per-field LLM-as-a-Judge because it is forced to allocate verification tokens for each field and thus attend to them all.

Combining per-document and per-field LLM-as-a-Judge mitigates their respective biases, but the broader shortcomings S3-S5 remain challenges that necessitate trade-offs.
As an analogy, consider asking human employees to verify structured outputs.  For a senior employee, you might specify only the main goal, expecting they will intuit a global judgement by leveraging their experience.  For a junior employee, you might specify more tips/steps, expecting they will process low-level details with higher-fidelity but may fare worse at allocating focus and grasping the bigger picture.  For more reliable assessment, you might utilize the \emph{wisdom of the crowd}, pragmatically limiting your requests to a few coworkers \citep{page2007difference, kuncheva2003measures}. 

CONSTRUCT similarly aims to produce reliable trustworthiness scores with few verifier LLM calls, by ensuring they are \textbf{diverse} yet \textbf{individually effective} (so that we do not need latency-impacting coordination across LLM calls) \citep{hu2025dipper}.  
For diversity, our different LLM calls receive different: tasks, scoring mechanisms/formats, and chain-of-thought criteria to guide reasoning \citep{kojima2022large}.
We improve detector coverage by splitting assessments across different LLM calls rather than forcing one call to do too much.  
To get more out of few verifier calls, some of our LLM calls do simultaneously contribute to per-document and per-field scoring in a manner that delicately circumvents shortcoming S4.  Ultimately, CONSTRUCT's verifier LLM calls represent a delicate balance between: information-processing fidelity vs.\ diversity vs.\ quality/redundancy.  We comprehensively explored this design space when prototyping CONSTRUCT, settling on certain trade-offs that proved best in the enterprise workloads where our method is deployed.

\subsection{How CONSTRUCT works}
\label{sec:algo}

CONSTRUCT issues several verifier LLM calls that yield two types of intermediate scores: overall document-level scores $S_{\text{doc}}(x, \hat{y}) \in [0, 1]$ and field-level scores $S_i(x, \hat{y}_i) \in [0, 1]$. We produce several instances of each intermediate score type and then aggregate  intermediate scores into final per-document and per-field trustworthiness scores. 
To limit scoring \textbf{latency}, verifier LLM calls are made in parallel with no sequential dependencies between them. CONSTRUCT additionally employs a time-out mechanism that ignores the results of verifier LLM calls that did not return results within a certain (adaptively chosen) time-limit.  To reduce CONSTRUCT's latency/cost, simply use a faster/cheaper verifier LLM. 


Each call to the verifier uses a different prompt-template with distinct instructions for this LLM.  CONSTRUCT makes 5 LLM calls in total using the templates from Appendix \ref{app:prompt-template}.  There are three types of templates: 
\\[0.5em]
\textbf{Type 1: Document-level scoring.}
    These templates assess the full structured output, eliciting a score  $S_{\text{doc}}(x, \hat{y})$ reflecting the verifier’s certainty that the entire structured output is correct, complete, and factually accurate.
    The verifier is instructed to jointly consider all fields in \( \hat{y} \) and penalize: incorrect values including  mistakenly missing/null values, unverifiable claims, or inconsistencies.  
    \\[0.5em]
\textbf{Type 2: Field-level scoring.}  
    These templates elicit individual scores for every top-level field $\{S_i(x, \hat{y}_i)\}_{i=1}^n$ via a single LLM call that uses the Structured Outputs capability to organize these scores.  The verifier is instructed to evaluate each field for correctness, completeness, and verifiability.
  \\[0.5em]
\textbf{Type 3: Document-level scoring + indirect field-level scoring.}  
    These templates ask the verifier LLM to first determine which fields seem incorrect, incomplete, or uncertain (without producing field-level scores), and subsequently elicit a score assessing  overall confidence in the full structured output: $S_{\text{doc}}(x, \hat{y})$.  Let $\mathcal{I} \subseteq \{1, \dots, n\}$ denote the list of fields flagged by the verifier. We map these categorical judgments to numeric field-level scores via:
    \[
    S_i(x, \hat{y}_i) =
    \begin{cases}
    \alpha, & \text{if } i \in \mathcal{I}, \\
    \beta,  & \text{otherwise},
    \end{cases}
    \quad \text{with } 0 \le \alpha \ll \beta \le 1.
    \]
Our implementation of CONSTRUCT simply fixes $ \alpha = 0.1$ and $\beta = 0.9$.
The intermediate flagging of incorrect fields serves as a \emph{structured reasoning} technique \citep{zhouleast} to force the verifier model to log information that we know is critical to document-level scoring. Beyond often improving the document-level scores, this directed reasoning serves a dual purpose of providing field-level scoring from a binary classification perspective.  If the verifier LLM API offers token-probabilities, one could use them in place of $\alpha, \beta$, but we found these probabilities too overconfident and noisy to be effective.
\\[0.5em]


\noindent \textbf{Score Aggregation.}  After producing several intermediate document-level and field-level scores from the aforementioned LLM calls, we apply basic (hyperparameter-free) operations to aggregate these into per-document and per-field trustworthiness scores.  We first 
produce additional document-level intermediate scores by taking a \emph{harmonic average} of the intermediate field-level scores produced from Type 2 templates:
    \[
    S_{\text{doc}}(x, \hat{y}) = \frac{n}{\sum_{i=1}^{n} \frac{1}{S_i(x, \hat{y}_i)}}
    \]
Behaving as a soft-minimum over scores $S_i \in [0,1]$, harmonic averaging provides an effective way to aggregate estimated signals in global detection settings where overall failure is defined by the union of component-level failure events \citep{wilson2019harmonic} -- i.e., an output is incorrect if any field within it is incorrect.
Finally to produce the per-document trustworthiness score, we take a simple arithmetic average of all intermediate document-level scores $S_{\text{doc}}(x, \hat{y})$.
To produce the per-field trustworthiness score, we take a simple arithmetic average of all intermediate field-level scores $S_i(x, \hat{y}_i)$.

\section{A High-Quality Structured Outputs Benchmark}

While benchmarking Structured Outputs from leading LLMs, we found \textbf{major errors} in the provided ``ground-truth" outputs across existing popular benchmark datasets (detailed in Appendix \ref{app:existing-benchmark-issues}).  Noisy labels have been observed in other benchmark datasets \citep{northcutt-noisy-labels, kuan2022model}, but we found the situation far more dire in every Structured Outputs dataset that we studied. 
To enable more reliable LLM evaluation, we introduce four Structured Outputs benchmarks with verified high-quality ground-truth outputs. We conducted extensive error analysis with frontier models and manual inspection to ensure the ground-truth outputs provided with each dataset are significantly more reliable than existing public benchmarks.

Each of our benchmark datasets contains many examples from a different use-case, and is formatted for straightforward application of LLM Structured Outputs. Each example comes with: a LLM prompt, input to process (i.e. text document), JSON schema describing how the output should be structured (composed of various fields whose values must be of a declared data type). Solely for evaluation-purposes, there is also a ground-truth structured output which any LLM output can easily be compared against (to determine whether the LLM handled this example correctly or not).
Our benchmarks involve diverse Structured Outputs, spanning fields with nested JSON output, lists, numbers, categories, and other text.  Below we briefly describe each dataset in our benchmark, showing specific examples from each dataset in Appendix \ref{app:benchmark-input-output}.

\subsection{Dataset: Financial Entities Extraction}

This benchmark is a refined version of the original FIRE dataset from \citet{Hassan2023fire}, with improved consistency, clarity, and accuracy in ground-truth annotations. Each sample contains a text document from business or financial news, reports, or corporate filings. The task is to extract  specific financial and contextual entities mentioned in the text.

We refined the original dataset by cleaning up annotation errors, removing ambiguous fields, and merging the overlapping fields into a unified category.
From the original dataset, we removed several ambiguous or overlapping fields: \textsf{Action}, \textsf{BusinessUnit}, \textsf{Designation}, \textsf{FinancialEntity}, and \textsf{Sector}. In addition, the original \textsf{GeopoliticalEntity} and \textsf{Location} fields have been merged into a unified \textsf{Location} field to resolve inconsistencies in how geographic mentions were annotated. Our resulting benchmark has seven better scoped fields for extraction: \textsf{Company}, \textsf{Date}, \textsf{Location}, \textsf{Money}, \textsf{Person}, \textsf{Product}, and \textsf{Quantity}.

\subsection{Dataset: PII Extraction}
This benchmark is a refined version of the original PII Entity Recognition dataset from \citet{marie_stephen_leo_2024_12327267}, with several annotation errors fixed. Each sample in this benchmark contains various types of personal information embedded within a document. The task is to identify, extract, and categorize different forms of PII in the document (there are 56 PII categories including: \textsf{age}, \textsf{credit card}, \textsf{city}, \textsf{street}, \textsf{IPV6}, ...).

\subsection{Dataset: Insurance Claims Extraction}

This benchmark contains documents we wrote to represent insurance claims (such as scanned forms, intake notes, or customer-submitted reports). The task is to extract information about each claim, including policy details, insured objects, and incident descriptions.
Models should produce a nested JSON output with four main fields, each containing subfields that capture key elements such as: administrative identifiers ( \textsf{claim ID} and  \textsf{report date}), policy information (\textsf{policy number},  \textsf{holder name},  \textsf{coverage type},  \textsf{policy period}), details of insured properties (\textsf{type},  \textsf{year built},  \textsf{address},  \textsf{estimated value}), and incident-specific information (\textsf{incident type},  \textsf{location},  \textsf{damage estimates},  \textsf{police report references}).

\subsection{Dataset: Data Table Analysis}

This benchmark contains structured data tables we created, and the task is to output key metadata about each table. Each sample contains a table in a CSV formatted string, and models need to extract information such as the number of rows and columns, column types, minimum and maximum values, and other statistics. For each sample, models should output a total of seven fields, which include a mix of string, numeric, and dictionary (object) types.  Getting some of these output fields right requires complex information aggregation and reasoning beyond mere extraction.

\section{Benchmarking LLM Accuracy}

For each example in our datasets, we generated Structured Outputs using each of the following models:  OpenAI’s GPT-5, GPT-4.1-mini, Google's Gemini 3 Pro, Gemini 2.5 Pro, and Gemini 2.5 Flash (run with its default thinking mode enabled) \citep{singh2025openai,openai2025gpt41,comanici2025gemini,deepmind_gemini3pro_model_card_2025}.  These are currently the LLMs most commonly used for Structured Output applications by enterprises that we surveyed.
For each benchmark dataset, we evaluate LLM-generation performance using two metrics:
\begin{itemize}
\item \textbf{Field Accuracy}: measures the proportion of individual fields that are extracted correctly (across all fields from all samples' outputs).
\item \textbf{Document Accuracy}: measures the proportion of samples/documents for which every field is correct in the LLM's output (a sample/document is considered incorrectly-handled by the LLM if even a single field is incorrect).
\end{itemize}

\begin{table}[tb!]
\small
\centering
\setlength{\tabcolsep}{4pt}
\begin{tabular}{lccccc}
\toprule
 & GPT-4.1-mini & GPT-5 & Gemini-2.5-Pro & Gemini-2.5-Flash & Gemini-3-Pro \\
 \midrule
 \rowcolor{gray!20}
\multicolumn{6}{l}{\textbf{Financial Entities Extraction}} \\
Field Accuracy & 0.922 & \textbf{0.949} & 0.887 & 0.919 & 0.935 \\
Document Accuracy  & 0.580 & \textbf{0.700} & 0.422 & 0.557 & 0.624 \\
 \rowcolor{gray!20}
\multicolumn{6}{l}{\textbf{PII Extraction}} \\
Field Accuracy & 0.966 & \textbf{0.979} & 0.972 & 0.973 & \textbf{0.979} \\
Document Accuracy & 0.260 & \textbf{0.460} & 0.300 & 0.330 & 0.440 \\
 \rowcolor{gray!20}
\multicolumn{6}{l}{\textbf{Insurance Claims Extraction}} \\
Field Accuracy & 0.750 & 0.767 & \textbf{0.775} & 0.758 & \textbf{0.775} \\
Document Accuracy  & 0.333 & 0.300 & \textbf{0.400} & 0.300 & 0.300 \\
 \rowcolor{gray!20}
\multicolumn{6}{l}{\textbf{Data Table Analysis}} \\
Field Accuracy & 0.863 & 0.956 & 0.944 & 0.829 & \textbf{0.964} \\
Document Accuracy  & 0.450 & 0.760 & 0.720 & 0.280 & \textbf{0.770} \\
\bottomrule
\end{tabular}
\caption{Accuracy of various LLMs in each of our Structured Outputs benchmarks.}
\label{tab:acc}
\end{table}

Table \ref{tab:acc} shows that no one model dominates all of the benchmarks. For some tasks, the smaller-tier models can roughly match the accuracy of their larger counterparts.  Recurring errors we observe in the LLM outputs include: getting confused when there are multiple similar instances of a type of entity (e.g. dates in a document), wrongly associating entities with concepts that are not actually related to a target field, mis-formatting extractions or mis-extracting the right span, overlooking information, and arriving at the wrong value for outputs that require reasoning rather than mere extraction.  Appendix \ref{app:detectederrors} shows example errors.  Gemini 3 Pro cost us 1.5x more to run than GPT-5 for these benchmarks (using more thinking tokens). Over these benchmarks, GPT-4.1-mini was cheaper than Gemini-2.5-Flash in its default thinking-enabled mode (only 20\% of the cost), and it was lower latency on average (GPT-4.1-mini's P50 latency was 65\% of Gemini-2.5-Flash's).

\section{Benchmarking Trustworthiness Scores}

Next we evaluate various per-document and per-field trustworthiness scores, and which scores best detect errors in the outputs from these different LLMs.

\subsection{Setup}
We compare CONSTRUCT against alternative scoring methods described in Sections \ref{sec:relatedwork} and \ref{sec:method}.
While the LLM outputs are generated using a variety of models in our experiments, all trust scoring of these LLM outputs is performed using CONSTRUCT powered by GPT-4.1-mini. For consistency, all LLM-as-a-Judge evaluations also use GPT-4.1-mini.  At the time we ran our benchmarks, GPT-4.1-mini is a preferred choice for real-time LLM verification amongst enterprises we surveyed (thanks to its low latency/cost, solid instruction-adherence and world-knowledge, mature deployment infrastructure via Azure/OpenAI APIs, and higher accuracy for many real-world tasks than comparably priced LLMs). 
For results where the output is generated by GPT-5 or Gemini-3-Pro, Token-Probability-based scoring is omitted because these models did not expose token-level log probabilities on the date we ran our benchmarks. 

We evaluate the score-based error detectors by their \emph{Area Under the Receiver Operating Characteristic Curve} (AUROC), which measures how well their scores rank incorrect LLM outputs above correct ones. Higher AUROC means a detector both catches more incorrect outputs (higher recall) and is more accurate when flagging them (higher precision).  When evaluating per-document scores, a LLM output is considered incorrect if any of its fields is wrong.  When evaluating per-field scores, the correctness of each field within a LLM output is considered independently of the other fields/outputs.

Appendix \ref{app:moreresults} presents additional evaluations of scoring methods under different performance metrics.  Appendix \ref{app:experimentdetails} provides additional details about our experimental setup.

\subsection{Results}

\begin{figure*}[tbh!]
\centering
  \includegraphics[width=1.06\linewidth]{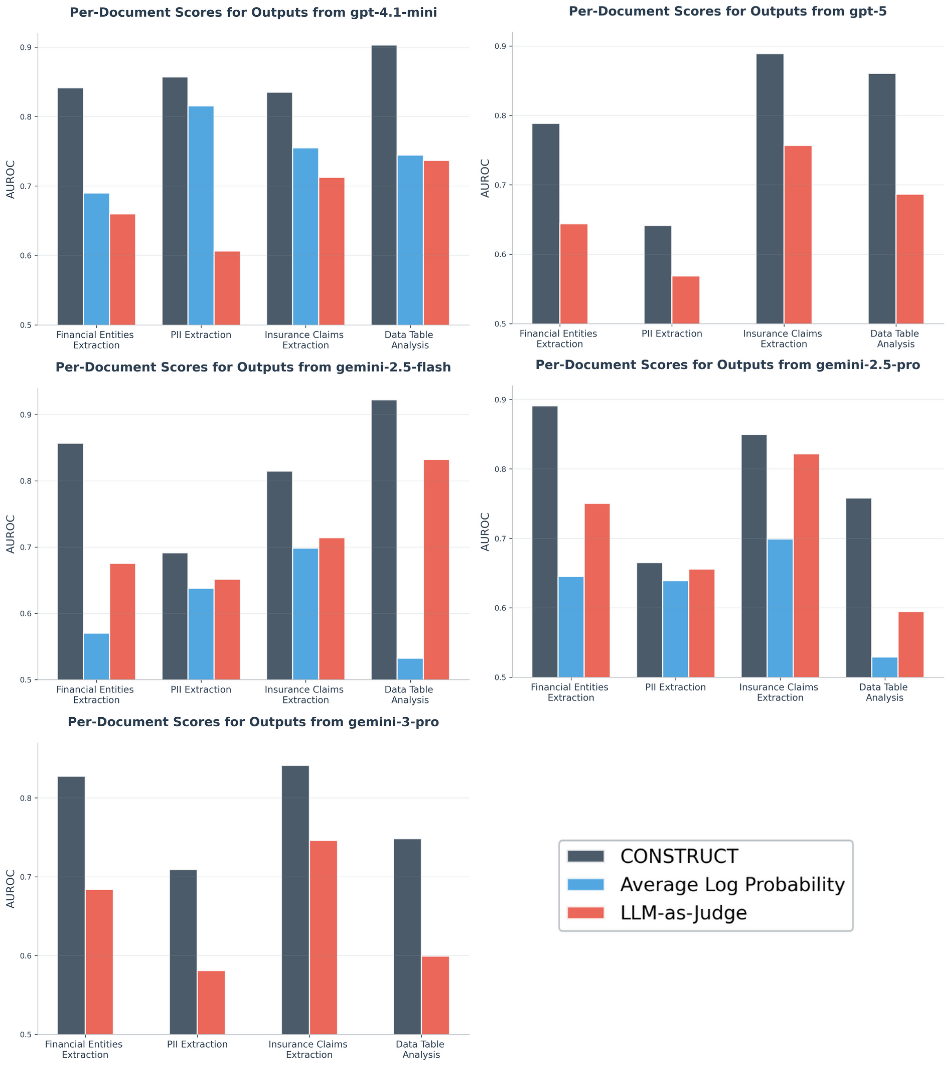}
  \caption{How effectively do different \textbf{Per-Document} scoring techniques detect \emph{incorrect} LLM Structured Outputs. Here a document's LLM output is deemed incorrect if any of its fields are wrong, the subtitle of each graph indicates which model produced the outputs, and detection effectiveness is measured via \textbf{AUROC} (higher is better).}
  \label{fig:auroc_doc}
\end{figure*}

\begin{figure*}[tbh!]
\centering
  \includegraphics[width=1.06\linewidth]{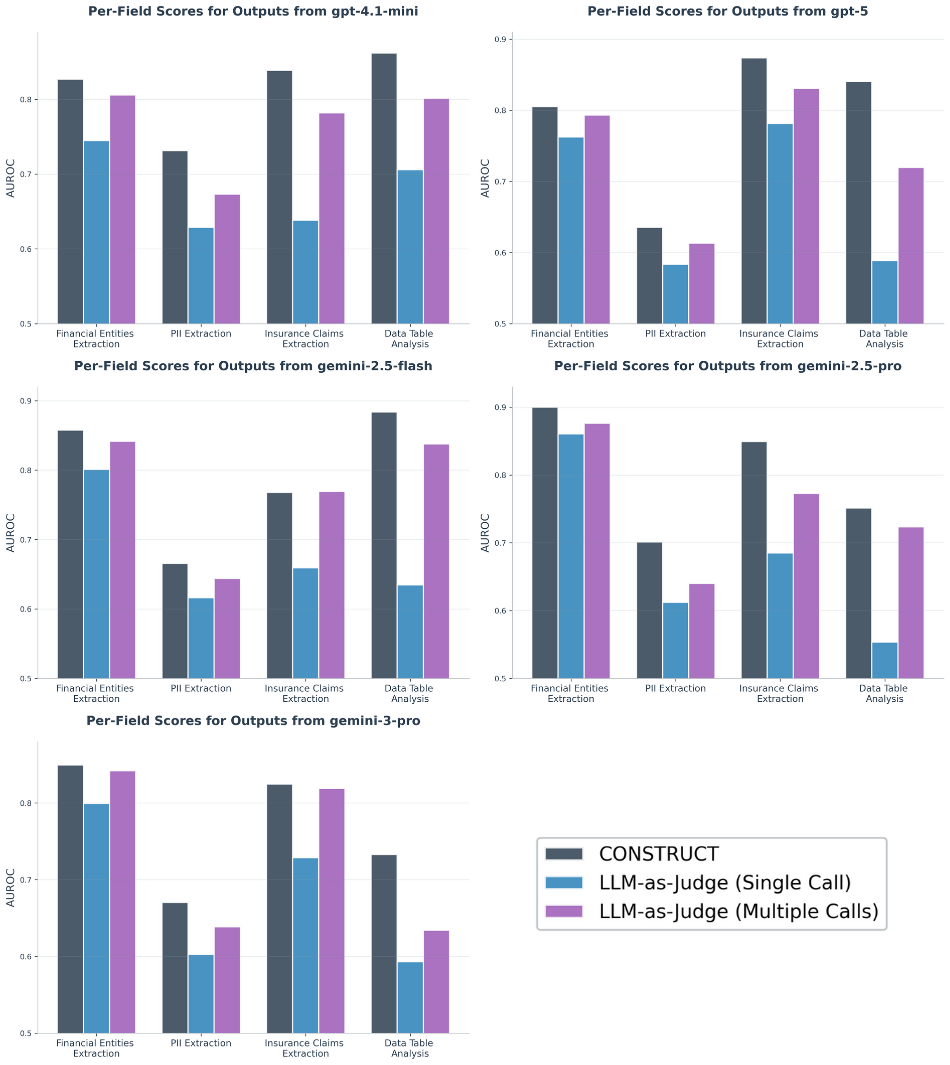}
\caption{How effectively (in terms of \textbf{AUROC}) do different \textbf{Per-Field} scoring techniques detect \emph{individual fields} that are erroneous within LLM Structured Outputs.}
  \label{fig:auroc_field}
\end{figure*}

Figure \ref{fig:auroc_doc} shows the performance of different Per-Document trust scores across our benchmark (also see Figures \ref{fig:precision_doc} and   \ref{fig:confidencegap_doc} in the Appendix).
Across all five models studied here, CONSTRUCT's per-document trust scores achieve the highest AUROC on every benchmark and by a substantial margin. This detector consistently exhibits the best precision/recall for catching incorrect Structured Outputs. LLM-as-a-Judge delivers only moderate performance, while scoring based on the average token log probability is inconsistently effective, varying widely across models.

Figure  \ref{fig:auroc_field} shows the performance of different Per-Field trust scores across our benchmark (also see Figures \ref{fig:precision_field} and \ref{fig:confidencegap_field} in the Appendix).  For per-field scoring to detect which specific output fields are erroneous, the multiple-call LLM-as-a-Judge method consistently outperforms the single-call version (but at the cost of making many more LLM calls). CONSTRUCT consistently outperforms both LLM-as-a-Judge baselines while consuming less tokens/compute than the multiple-call approach. This efficiency advantage grows even more significant as the number of fields increases, since methods that require a separate LLM-Judge call per field will scale poorly to large Structured Outputs.

\section{Discussion}
\label{sec:discussion}

As enterprise LLM applications increasingly rely on Structured Outputs, effective trust assessment grows increasingly important. Our comprehensive benchmarks revealed that basic methods like token-probabilities and LLM-as-a-Judge scores are suboptimal for the reliable error-detection required in these workflows. CONSTRUCT provides a scalable, real-time solution: trustworthiness scores that accurately flag incorrect outputs, pinpointing which fields may be wrong (via per-field scores) and why (via automated explanations).

Beyond its favorable precision/recall and latency/cost, our method enjoys the following advantages:
\begin{itemize}
\item CONSTRUCT is straightforward to implement and generalizes to arbitrary Structured Outputs applications and LLM models (including black-box LLM APIs that do not even expose token probabilities).
\item CONSTRUCT does not require dataset preparation/labeling work nor custom model training/serving infrastructure, and works out-of-the-box.
\item CONSTRUCT scoring immediately improves whenever better LLMs are released and adopted as its base model (unlike custom trained models which will struggle to detect errors from future advanced LLMs, unless they are retrained to be significantly better, e.g., with more annotated data).
\end{itemize}

These advantages drove several enterprises we worked with to integrate CONSTRUCT into high-stakes LLM applications, and multiple finance/insurance companies reported that this integration was critical to successfully productionizing their applications. 

\clearpage
\bibliography{structured_outputs}


\newpage
\appendix



\section{Example LLM inputs and outputs from our Structured Outputs Benchmark} 
\label{app:benchmark-input-output}

For each example in our benchmarks, the LLM receives: a \emph{system message} (i.e., system prompt/instructions) specifying the overall task and how to behave, a \emph{user message} containing the specific document/information upon which its Structured Output should be based, and a desired  \emph{schema} its Structured Output must adhere to.  The system message and schema remain fixed across all examples from one dataset.

\noindent Refer to our Structured Outputs benchmark code/data for more details: \\ \url{https://github.com/cleanlab/structured-output-benchmark}

\subsection{Dataset: Financial Entities Extraction}


\begin{tcolorbox}[colback=gray!10,colframe=black!60!black,
                  title=\textbf{System Instructions}, boxrule=0.5mm, arc=3mm, breakable]
\begin{BreakVerbatim}
Identify and extract entities from the following financial news text into the following categories:

Entity 1: Company

- Definition: Denotes the official or unofficial name of a registered company or a brand.
- Example entities: \{Apple Inc.; Uber; Bank of America\}

Entity 2: Date

- Definition: Represents a specific time period, whether explicitly mentioned (e.g., "year ended March 2020") or implicitly referred to (e.g., "last month"), in the past, present, or future.
- Example entities: \{June 2nd, 2010; quarter ended 2021; last week; prior year; Wednesday\}

Entity 3: Location

- Definition: Represents geographical locations, such as political regions, countries, states, cities, roads, or any other location, even when used as adjectives.
- Example entities: \{California; Paris; 1280 W 12th Blvd; Americas; Europe\}

Entity 4: Money

- Definition: Denotes a monetary value expressed in any world currency, including digital currencies.
- Example entities: \{\$76.3 million; \$4 Bn; Rs 33.80 crore; 1.2 BTC\}

Entity 5: Person

- Definition: Represents the name of an individual.
- Example entities: \{Meg Whitman; Mr. Baker; Warren Buffet\}

Entity 6: Product

- Definition: Refers to any physical object or service manufactured or provided by a company to consumers, excluding references to businesses or sectors within the financial context.
- Example entities: \{iPhone; Tesla model X; cloud services; Microsoft Windows 10; laptops; medical equipment; computer software; online classes; eye surgery\}

Entity 7: Quantity

- Definition: Represents any numeric value that is not categorized as Money, such as percentages, numbers, measurements (e.g., weight, length), or other similar quantities. Note that unit of measurements are also part of the entity.
- Example entities: \{15%; 25,000 units; 2.75in; 100 tons\}

For each category:

- Extract all relevant entities as a list of strings, preserving the wording from the text
- Use None if no entities are found in that category
- Only extract entities that are explicitly mentioned in the text itself, do not make inferences or reason about what entities might be implied based on URLs, domain names, or other indirect references
- Extract individual items rather than compound or ranged entities (e.g., if a range or compound entity is mentioned, extract each individual item separately)

Return the extracted information as a JSON object with all categories included, using None for cases where no entities are found.
\end{BreakVerbatim}
\end{tcolorbox}

\begin{tcolorbox}[colback=gray!10,colframe=black!60!black,
                  title=\textbf{Sample User Message}, boxrule=0.5mm, arc=3mm]
\begin{BreakVerbatim}
GSK rejected \$68 billion Unilever bid for consumer division
\end{BreakVerbatim}
\end{tcolorbox}

\begin{tcolorbox}[colback=gray!10,colframe=black!60!black,
                  title=\textbf{Sample Output}, boxrule=0.5mm, arc=3mm]
\begin{verbatim}
{
  "Company": ["GSK", "Unilever"],
  "Date": None,
  "Location": None,
  "Money": ["$68 billion"],
  "Person": None,
  "Product": None,
  "Quantity": None
}
\end{verbatim}
\end{tcolorbox}

\subsection{Dataset: PII Extraction}

\begin{tcolorbox}[colback=gray!10,colframe=black!60!black,
                  title=\textbf{System Instructions}, boxrule=0.5mm, arc=3mm, breakable]
\begin{BreakVerbatim}
Your task is to extract structured information about PII entities from the text provided by the user.
For each field in the response format, extract the corresponding PII entity if it exists in the text. If a particular PII entity is not present in the text, set that field to null.
Return the complete structured response with all fields, setting missing entities to null.
\end{BreakVerbatim}
\end{tcolorbox}

\begin{tcolorbox}[colback=gray!10,colframe=black!60!black,
                  title=\textbf{Sample User Message}, boxrule=0.5mm, arc=3mm]
\begin{BreakVerbatim}
Your course fee payment receipt is attached. A charge of ISK 342,754.57 has been successfully transacted from your credit card 3534282572218237. Halvorson Inc thanks you for choosing to invest in lifelong learning.
\end{BreakVerbatim}
\end{tcolorbox}

\begin{tcolorbox}[colback=gray!10,colframe=black!60!black,
                  title=\textbf{Sample Output}, boxrule=0.5mm, arc=3mm, breakable]
\begin{verbatim}
{
  "ACCOUNTNAME": None,
  "ACCOUNTNUMBER": None,
  "AGE": None,
  "AMOUNT": "342,754.57",
  "BIC": None,
  "BITCOINADDRESS": None,
  "BUILDINGNUMBER": None,
  "CITY": None,
  "COMPANYNAME": "Halvorson Inc",
  "COUNTY": None,
  "CREDITCARDCVV": None,
  "CREDITCARDISSUER": None,
  "CREDITCARDNUMBER": "3534282572218237",
  "CURRENCY": None,
  "CURRENCYCODE": "ISK",
  "CURRENCYNAME": None,
  "CURRENCYSYMBOL": None,
  "DATE": None,
  "DOB": None,
  "EMAIL": None,
  "ETHEREUMADDRESS": None,
  "EYECOLOR": None,
  "FIRSTNAME": None,
  "GENDER": None,
  "HEIGHT": None,
  "IBAN": None,
  "IP": None,
  "IPV4": None,
  "IPV6": None,
  "JOBAREA": None,
  "JOBTITLE": None,
  "JOBTYPE": None,
  "LASTNAME": None,
  "LITECOINADDRESS": None,
  "MAC": None,
  "MASKEDNUMBER": None,
  "MIDDLENAME": None,
  "NEARBYGPSCOORDINATE": None,
  "ORDINALDIRECTION": None,
  "PASSWORD": None,
  "PHONEIMEI": None,
  "PHONENUMBER": None,
  "PIN": None,
  "PREFIX": None,
  "SECONDARYADDRESS": None,
  "SEX": None,
  "SSN": None,
  "STATE": None,
  "STREET": None,
  "TIME": None,
  "URL": None,
  "USERAGENT": None,
  "USERNAME": None,
  "VEHICLEVIN": None,
  "VEHICLEVRM": None,
  "ZIPCODE": None
}
\end{verbatim}

\end{tcolorbox}

\subsection{Dataset: Insurance Claims Extraction}

\begin{tcolorbox}[colback=gray!10,colframe=black!60!black,
                  title=\textbf{System Instructions}, boxrule=0.5mm, arc=3mm]
\begin{BreakVerbatim}
You are an expert insurance claim processor. Extract structured information from insurance claim descriptions.
For dates, use YYYY-MM-DD format.
If a piece of information does not exist in the claim description, return null instead of making assumptions.
Be thorough in extracting all available information and categorize appropriately.
\end{BreakVerbatim}
\end{tcolorbox}

\begin{tcolorbox}[colback=gray!10,colframe=black!60!black,
                  title=\textbf{Sample User Message}, boxrule=0.5mm, arc=3mm, breakable]
\begin{BreakVerbatim}
Please extract the insurance claim information from this text:

Logged in user: James Smith.
Submission timestamp: 16-8-2025.
Submission id: CLM-405338.
Assigned policies: - POL-932318703 car insurance policy.
Item VIN70413800966102220 is a Honda CR-V, 2017 model.
Activated September 25th, 2024.
Valid for 1 year.

Incident description:
I driving yesterday down the highway when, near Harrison Avenue, the other car just came out of nowhere on the opposite side, and next thing I knew, bam-crashed into the front. Like, I tried to swerve, but there wasn't really anywhere to go, just those concrete dividers and the light pole on the right. My car took most of the hit. Estimate of damages is over \$37,000 according to the mechanic, but I was surprised hese numbers can get up there when I think the car is worth half that.
\end{BreakVerbatim}
\end{tcolorbox}

\begin{tcolorbox}[colback=gray!10,colframe=black!60!black,
                  title=\textbf{Sample Output}, boxrule=0.5mm, arc=3mm, breakable]
\begin{verbatim}
{
  "header": {
    "claim_id": "CLM-405338",
    "report_date": "2025-08-16",
    "incident_date": "2025-08-15",
    "reported_by": "James Smith",
    "channel": "Portal"
  },
  "policy_details": {
    "policy_number": "POL-932318703",
    "policyholder_name": "James Smith",
    "coverage_type": "Auto",
    "effective_date": "2024-09-25",
    "expiration_date": "2025-09-25"
  },
  "insured_objects": [
    {
      "object_id": "VIN70413800966102220",
      "object_type": "Vehicle",
      "make_model": "Honda CR-V",
      "year": 2017,
      "location_address": None,
      "estimated_value": None
    }
  ],
  "incident_description": {
    "incident_type": "head_on_collision",
    "location_type": "highway",
    "estimated_damage_amount": 37000,
    "police_report_number": None
  }
}
\end{verbatim}
\end{tcolorbox}

\subsection{Dataset: Data Table Analysis}

\begin{tcolorbox}[colback=gray!10,colframe=black!60!black,
                  title=\textbf{System Instructions}, boxrule=0.5mm, arc=3mm]
\begin{BreakVerbatim}
You are given a CSV-like string representation of a table (with header row, no index).
Extract a structured JSON object following the provided response format class. Do not guess: if a value does not exist or is not applicable, return null.
Count rows excluding the header. Infer each column type as 'str', 'int', or 'float'. For string columns, set min/max to null. If the 'Identifier' column is missing, set all Identifier-related fields to null. For null/None entries, set string columns to '', and numerical to None.
Return only the structured JSON object.
\end{BreakVerbatim}
\end{tcolorbox}

\begin{tcolorbox}[colback=gray!10,colframe=black!60!black,
                  title=\textbf{Sample User Message}, boxrule=0.5mm, arc=3mm]
\begin{BreakVerbatim}
Identifier,Beta,Gamma
YaYMXPIq,,-667.620811
xxOiAcMs,-707,2485.0,-940.618915
,849.0,930.809812
\end{BreakVerbatim}
\end{tcolorbox}

\begin{tcolorbox}[colback=gray!10,colframe=black!60!black,
                  title=\textbf{Sample Output}, boxrule=0.5mm, arc=3mm]
\begin{verbatim}
{
  "num_rows": 3,
  "num_columns": 3,
  "column_types": {"Identifier": "str", "Beta": "int", "Gamma": "float"},
  "column_max": {"Identifier": None, "Beta": 2485.0, "Gamma": 930.809812},
  "column_min": {"Identifier": None, "Beta": 849.0,"Gamma": -940.618915},
  "identifier_first": "YaYMXPIq",
  "identifier_last": "",
  "identifier_shortest": ""
}
\end{verbatim}
\end{tcolorbox}

\newpage
\section{Example LLM Errors Detected using CONSTRUCT Trustworthiness Scoring}
\label{app:detectederrors}

Here we showcase example errors that LLMs made in this benchmark. All examples in this section are from  outputs generated by the GPT-4.1-mini LLM. For brevity, we omit the specified output schema and the system instructions (see Appendix \ref{app:benchmark-input-output}).

\subsection{Financial Entities Extraction Example}

\begin{quote}
\begin{lstlisting}[language=Python] 
User Input (see Appendix A for System Instructions): 
\end{lstlisting}
\vspace*{-0.5em}
\begin{lstlisting}[style=output]
Crestview could sell its stake in us to one or more third parties that may not be favorable to our stockholders.
\end{lstlisting}

\begin{lstlisting}[language=Python] 
Ground Truth Response:
\end{lstlisting}
\vspace*{-0.5em}
\begin{lstlisting}[style=output]
{
  'Company': ['Crestview'],
  'Date': None,
  'Location': None,
  'Money': None,
  'Person': None,
  'Product': None,
  'Quantity': None
}
\end{lstlisting}

\begin{lstlisting}[language=Python] 
LLM Response:
\end{lstlisting}
\vspace*{-0.5em}
\begin{lstlisting}[style=output]
{
  'Company': None,
  'Date': None,
  'Location': ['us'],
  'Money': None,
  'Person': None,
  'Product': None,
  'Quantity': None
}
\end{lstlisting}

\begin{lstlisting}[language=Python] 
CONSTRUCT Trustworthiness Score: 
\end{lstlisting}
\vspace*{-0.5em}
\begin{lstlisting}[style=output]
0.103577
\end{lstlisting}

\begin{lstlisting}[language=Python] 
Per-Field Trustworthiness Scores and Explanations:
\end{lstlisting}
\vspace*{-0.5em}
\begin{lstlisting}[style=output]
{
  'Company': {
    'explanation': 'The text mentions 'Crestview' which appears to be a company name, so it should be extracted under Company. The response has null for Company, which is incorrect.',
    'score': 0.05
  },
  'Date': {
    'explanation': 'No specific dates or time periods are mentioned in the text, so null is appropriate.',
    'score': 0.95
  },
  'Location': {
    'explanation': 'The response incorrectly classifies 'us' as a Location, but 'us' is a pronoun and not a geographical location.',
    'score': 0.1875
  },
  'Money': {
    'explanation': 'No monetary values are mentioned in the text, so null is appropriate.',
    'score': 0.95
  },
  'Person': {
    'explanation': 'No individual person names are mentioned in the text, so null is appropriate.',
    'score': 0.95
  },
  'Product': {
    'explanation': 'No products or services are mentioned in the text, so null is appropriate.',
    'score': 0.95
  },
  'Quantity': {
    'explanation': 'No quantities such as percentages or numbers are mentioned in the text, so null is appropriate.',
    'score': 0.95
  }
}
\end{lstlisting}
\end{quote}

\subsection{PII Extraction Example}

\begin{quote}
\begin{lstlisting}[language=Python] 
User Input (see Appendix A for System Instructions): 
\end{lstlisting}
\vspace*{-0.5em}
\begin{lstlisting}[style=output]
Just wanted to give you a heads up that the sex education seminar has been moved NE Northeast, near the coords [-89.6132,167.3028]. See you there!
\end{lstlisting}

\begin{lstlisting}[language=Python] 
Ground Truth Response (some fields omitted for clarity):
\end{lstlisting}
\vspace*{-0.5em}
\begin{lstlisting}[style=output]
{
  'NEARBYGPSCOORDINATE': '[-89.6132,167.3028]',
  'ORDINALDIRECTION': 'Northeast',
  'SEX': None,
  ...
}
\end{lstlisting}

\begin{lstlisting}[language=Python] 
LLM Response (some fields omitted for clarity):
\end{lstlisting}
\vspace*{-0.5em}
\begin{lstlisting}[style=output]
{
  'NEARBYGPSCOORDINATE': '[-89.6132,167.3028]',
  'ORDINALDIRECTION': 'NE Northeast',
  'SEX': 'sex',
  ...
}
\end{lstlisting}

\begin{lstlisting}[language=Python] 
CONSTRUCT Trustworthiness Score: 
\end{lstlisting}
\vspace*{-0.5em}
\begin{lstlisting}[style=output]
0.411233
\end{lstlisting}

\begin{lstlisting}[language=Python] 
Per-Field Trustworthiness Scores and Explanations (some fields omitted):
\end{lstlisting}
\vspace*{-0.5em}
\begin{lstlisting}[style=output]
{
  'NEARBYGPSCOORDINATE': {
    'explanation': 'The input text contains coordinates [-89.6132,167.3028], which are correctly extracted.',
    'score': 0.95
  },
  'ORDINALDIRECTION': {
    'explanation': 'The field is extracted as \'NE Northeast\', which is redundant and inconsistent; the text only indicates a direction \'NE Northeast\' as a phrase, but this is not a valid ordinal direction value.',
    'score': 0.75
  },
  'SEX': {
    'explanation': 'The word 'sex' appears in 'sex education seminar' but it is not a PII sex/gender value; however, the response extracted 'sex' as the value which is incorrect because it is part of a phrase, not a standalone sex identifier.',
    'score': 0.05
  },
  ...
}
\end{lstlisting}
\end{quote}

\subsection{Insurance Claims Extraction Example}

\begin{quote}
\begin{lstlisting}[language=Python] 
User Input (see Appendix A for System Instructions): 
\end{lstlisting}
\vspace*{-0.5em}
\begin{lstlisting}[style=output]
Please extract the insurance claim information from this text:

Hi there, this is David Davis. I'm calling because, well, I was told to call my insurance company and file something about what happened today at my store. I'm still a little shook up, to be honest-the whole thing was just, I don't know, out of nowhere. Anyway, I don't really know what to say, but when I first called in, the lady I spoke to gave me a claim number-CLM-531517, I think she said-so I'm hoping that's the right one. She told me to keep it handy, so I've got it written down here on a sticky note next to my computer.

So today-I mean, it's the same day I'm calling you, which is August 22nd 2025, right? Yeah, Thursday. All the mornings blend together for me, but I keep track when things like this happen. Anyway, there was a customer, I think her name was Mrs. Patterson (she comes in a lot for our cinnamon bread), but I'm not totally sure-it's all a blur since the paramedics showed up. She slipped right in the front by the big window, near the display with the plants and the birthday cards. Someone had spilled water, I guess-though I swear we try to keep that area dry, especially after we mop in the morning. It wasn't even raining, so I don't know where the water came from, maybe from the flower buckets.

She went down pretty hard. I called 911 right away. There was surprisingly a police officer nearby-I think he was getting his coffee at the corner-so he was at the shop before the ambulance even got there. I think the police did a report, or at least that's what he said. All the paperwork was overwhelming, but I believe the number he gave me was PR-20250822-6127. I'm not really sure how these things go but I hope that helps.

It's been chaotic, honestly. While all that was going on, the bread delivery showed up and my dog kept barking from the back room, totally unrelated, but it was just a lot happening at once. I tried to check if she was okay but I don't know first aid or anything, and I didn't want to move her.

Now everyone keeps asking if she's going to try to sue or something, which is making me pretty nervous. A couple of my employees were there-Lisa and Jamie-and they're a bit shaken too. We closed for the afternoon to clean up and everyone's just kind of jumpy now. That's probably not important for the claim but I thought you should know.

Just to be clear, I'm reporting this on the phone right now (they said that would be fine), and the incident happened earlier today, before noon. The officer said he'd send a copy of the report to whoever needed it-insurance or lawyers, maybe.
\end{lstlisting}

\begin{lstlisting}[language=Python] 
Ground Truth Response:
\end{lstlisting}
\vspace*{-0.5em}
\begin{lstlisting}[style=output]
{
  'header': {
    'channel': 'Phone',
    'claim_id': 'CLM-531517',
    'incident_date': '2025-08-22',
    'report_date': '2025-08-22',
    'reported_by': 'David Davis'
  },
  'incident_description': {
    'estimated_damage_amount': None,
    'incident_type': 'property_injury',
    'location_type': 'commercial_property',
    'police_report_number': 'PR-20250822-6127'
  },
  'insured_objects': None,
  'policy_details': None
}
{
  'header': {
    'channel': 'Phone',
    'claim_id': 'CLM-531517',
    'incident_date': '2025-08-22',
    'report_date': '2025-08-22',
    'reported_by': 'David Davis'
  },
  'incident_description': {
    'estimated_damage_amount': None,
    'incident_type': 'property_injury',
    'location_type': 'commercial_property',
    'police_report_number': 'PR-20250822-6127'
  },
  'insured_objects': None,
  'policy_details': None
}
\end{lstlisting}

\begin{lstlisting}[language=Python] 
LLM Response:
\end{lstlisting}
\vspace*{-0.5em}
\begin{lstlisting}[style=output]
{
  'header': {
    'channel': 'Phone',
    'claim_id': 'CLM-531517',
    'incident_date': '2025-08-22',
    'report_date': '2025-08-22',
    'reported_by': 'David Davis'
  },
  'incident_description': {
    'estimated_damage_amount': None,
    'incident_type': 'property_injury',
    'location_type': 'commercial_property',
    'police_report_number': 'PR-20250822-6127'
  },
  'insured_objects': [
    {
      'estimated_value': None,
      'location_address': None,
      'make_model': None,
      'object_id': 'OBJ-000001',
      'object_type': 'Other',
      'year': None
    }
  ],
  'policy_details': None
}
\end{lstlisting}

\begin{lstlisting}[language=Python] 
CONSTRUCT Trustworthiness Score: 
\end{lstlisting}
\vspace*{-0.5em}
\begin{lstlisting}[style=output]
0.640920
\end{lstlisting}

\begin{lstlisting}[language=Python] 
Per-Field Trustworthiness Scores and Explanations:
\end{lstlisting}
\vspace*{-0.5em}
\begin{lstlisting}[style=output]
{
  'header': {
    'explanation': 'The claim ID CLM-531517 matches the claim number given by the user. The report date and incident date are both August 22, 2025, as stated in the text. The reporter is David Davis, and the claim was reported by phone as indicated by the user calling in. All these details are clearly stated and consistent.',
    'score': 0.95
  },
  'incident_description': {
    'explanation': 'The incident is a slip and fall at a commercial property (the store). The police report number PR-20250822-6127 is mentioned. No estimated damage amount is given. These details are consistent with the text and correctly categorized.',
    'score': 0.95
  },
  'insured_objects': {
    'explanation': 'The claim description does not specify any insured objects with identifiers, make/model, year, address, or value. The assistant included a generic object with ID OBJ-000001 and type 'Other' but no further details. This is an assumption and not supported by the text, so this is likely incorrect or at best uncertain.',
    'score': 0.25
  },
  'policy_details': {
    'explanation': 'No policy details such as policy number, policyholder name, coverage type, or policy dates are mentioned in the claim description. Returning null is appropriate.',
    'score': 0.95
  }
}
\end{lstlisting}
\end{quote}

\subsection{Data Table Analysis Example}

\begin{quote}
\begin{lstlisting}[language=Python] 
User Input (see Appendix A for System Instructions): 
\end{lstlisting}
\vspace*{-0.5em}
\begin{lstlisting}[style=output]
Alpha,Identifier
kWIrwaEqeJW,rkPiKnqE-367
,
\end{lstlisting}

\begin{lstlisting}[language=Python] 
Ground Truth Response:
\end{lstlisting}
\vspace*{-0.5em}
\begin{lstlisting}[style=output]
{
  'column_max': {'Alpha': None, 'Identifier': None},
  'column_min': {'Alpha': None, 'Identifier': None},
  'column_types': {'Alpha': 'str', 'Identifier': 'str'},
  'identifier_first': 'rkPiKnqE-367',
  'identifier_last": '',
  'identifier_shortest': '',
  'num_columns': 2,
  'num_rows': 2
}
\end{lstlisting}

\begin{lstlisting}[language=Python] 
LLM Response:
\end{lstlisting}
\vspace*{-0.5em}
\begin{lstlisting}[style=output]
{
  'column_max': {'Alpha': None, 'Identifier': 'rkPiKnqE-367'},
  'column_min': {'Alpha': None, 'Identifier': 'rkPiKnqE-367'},
  'column_types': {'Alpha': 'str', 'Identifier': 'str'},
  'identifier_first': 'rkPiKnqE-367',
  'identifier_last': '',
  'identifier_shortest': '',
  'num_columns': 2,
  'num_rows': 2
}
\end{lstlisting}

\begin{lstlisting}[language=Python] 
CONSTRUCT Trustworthiness Score: 
\end{lstlisting}
\vspace*{-0.5em}
\begin{lstlisting}[style=output]
0.418131
\end{lstlisting}

\begin{lstlisting}[language=Python] 
Per-Field Trustworthiness Scores and Explanations (some fields  omitted):
\end{lstlisting}
\vspace*{-0.5em}
\begin{lstlisting}[style=output]
{
  'column_max': {
    'explanation': 'For string columns, max should be null as per instructions, but 'Identifier' max is given as 'rkPiKnqE-367' which is incorrect; 'Alpha' max is correctly null.',
    'score': 0.05
  },
  'column_min': {
    'explanation': 'For string columns, min should be null as per instructions, but 'Identifier' min is given as 'rkPiKnqE-367' which is incorrect; 'Alpha' min is correctly null.',
    'score': 0.05
  },
  'column_types': {
    'explanation': 'Both columns contain string data (including empty string for missing), so both are correctly inferred as 'str'.',
    'score': 0.95
  },
  ...
}
\end{lstlisting}
\end{quote}

\newpage
\section{Issues with Existing Structured Outputs Benchmarks} \label{app:existing-benchmark-issues}

Given the importance of LLM Structured Outputs, a number of benchmarks have been introduced to evaluate model performance. In principle, comparing models should be straightforward: simply run them on these datasets. However, in practice, we found that many apparent ``errors’’ in the LLM outputs were actually mistakes in the ``ground-truth" output provided in the benchmark. 

To better understand this, we analyzed each dataset to identify potentially incorrect annotations. \emph{Every} public benchmark we reviewed contained a substantial amount of erroneous or inconsistent ground-truth data. Some level of noise is expected, but the degree of annotation problems we observed suggests that current public Structured Outputs benchmarks are too noisy to support reliable model accuracy assessments (and error-detection research depends even more critically on being able to reliably measure correctness).

Below are some specific example of errors we found in public benchmarks. Beyond the datasets listed in this section, we looked through many others as well, and also found them too noisy for reliable benchmarking.

\subsection{Financial Relation Extraction (FIRE)}

This dataset from \citet{Hassan2023fire} contains text samples from business or financial news, reports, or announcements. The task is to identify and extract specific financial and contextual entities mentioned in the text, including: \texttt{Action}, \texttt{BusinessUnit}, \texttt{Company}, \texttt{Date}, \texttt{Location}, \texttt{GeopoliticalEntity}, \texttt{FinancialEntity}, \texttt{Sector}, etc.

We found that this dataset is full of inconsistencies, where the ground truth contains certain extractions for some examples but not for other examples that are conceptually equivalent. It also has many ambiguities where one cannot judge whether an extracted output is truly correct or not based on the given instructions. Below are some examples of issues in this benchmark.

\subsubsection{Issue \#1: Ambiguity in defining what qualifies as a FinancialEntity}

There is inconsistency in how \texttt{FinancialEntity} is extracted across samples.
For instance, here are two samples in this dataset:

\begin{itemize}
\item \emph{``EFS acquired the assets and assumed certain liabilities of T - Chek.''}  \\
  $\rightarrow$ \texttt{FinancialEntity: [`assets']}
\item \emph{``On November 30 , 2010 , DXP acquired substantially all of the assets of D F Distributors , Inc. ( D F ).''} \\  
  $\rightarrow$ \texttt{FinancialEntity: None}
\end{itemize}

\noindent In the first sample, the term ``assets" is labeled as a \texttt{FinancialEntity}, but ``liabilities", another clear financial entity term in the same sentence, is not annotated. In the second sample, ``assets" again appears within the same semantics but is not labeled at all. 
This inconsistency introduces ambiguity, as it is unclear whether terms like assets, liabilities, and similar financial nouns should consistently be annotated as \texttt{FinancialEntity} or not.

\subsubsection{Issue \#2: Confusion between GeopoliticalEntity and Location labels}

There is inconsistency in how locations are categorized between \texttt{GeopoliticalEntity} and \texttt{Location}. Similar types of entities are labeled differently across samples, creating confusion over when a place should be considered a political entity vs a geographic location. 
For instance, here are two samples from this dataset:

\begin{itemize}
\item \emph{``Most storage operations are in north Louisiana and Oklahoma.''} \\  
  $\rightarrow$ \texttt{GeopoliticalEntity: [`Louisiana', `Oklahoma'], Location: None}
\item \emph{``Crossroads leases its headquarters , approximately 37,800 square feet of general office , laboratory , data center and administrative space in Austin , Texas.''} \\
  $\rightarrow$ \texttt{GeopoliticalEntity: None, Location: [`Austin', `Texas']}
\end{itemize}

\noindent In the first case, Louisiana and Oklahoma are treated as \texttt{GeopoliticalEntity}, while in the second, Texas, which is also a U.S. state, is labeled instead as a \texttt{Location}. This inconsistency makes it unclear when a geographic reference should be considered a political entity or a physical or regional location, leading to inconsistent annotations.

\subsubsection{Issue \#3: Incorrect annotations}

Here are a few samples of mislabeled data:

\begin{itemize}
\item ``In general, we do not compete directly with the major futures exchanges, such as CME, the Chicago Board Options Exchange, Eurex and Euronext, Liffe, and ICE.''  
  $\rightarrow$ \texttt{Designation: [`complete']} 

Complete is a verb and not a `Designation`, which typically includes extractions such as: CEO, President, Board of Directors, employees
  
\item ``We entered into an exclusive supply arrangement with R-Tech under which we granted R-Tech the exclusive right to manufacture and supply lubiprostone to meet its commercial and clinical requirement …''  
  $\rightarrow$ \texttt{Sector: [`clinical']} 

If "clinical" is extracted as a \texttt{Sector}, then ``commercial'' (another \texttt{Sector} in many of the examples) should also be included for consistency.

\end{itemize}

\subsection{PII Entity Recognition}

This dataset from \citet{marie_stephen_leo_2024_12327267} has text samples containing various types of personally identifying information (PII). The task is to extract different categories of PII from the text. Each sample may contains multiple types of PII that need to be identified and classified into specific categories including: names, credit card information, dates, etc. 
We found many of the ground truth annotations in this dataset are obviously wrong. Here are some examples:

\begin{itemize}
\item \emph{``Hope this message meets you well, Hertha. I have forwarded a set of past questions to your [Shirley.Goodwin@gmail.com](mailto:Shirley.Goodwin@gmail.com). Please answer and forward them to the same email before the examination date, 18/01/1960.''} \\
  $\rightarrow$ \texttt{DATE: "Not present", DOB: "18/01/1960"} 

18/01/1960 in the text is just a regular date (for an examination), not a date of birth.
  
\item \emph{``Hello Helene, schedule your upcoming physical therapy sessions by logging in with Opera/12.32 (X11; Linux i686; U; PL Presto/2.9.183 Version/10.00). Complete the process using 3115476202224400 on our secured portal.''} \\ 
  $\rightarrow$ \texttt{CREDITCARDNUMBER: "3115476202224400"} 

Given the context, the number appears in a technical or login context (plausibly a reference code or session token), not as an actual credit card number.

\item \emph{``Sign up for our course on handling learning disabilities at 40567. Secure payment options available at 3bCsN95v5LDqF7UFDo1wpYZtjX.''} \\ 
  $\rightarrow$ \texttt{ZIPCODE: "40567"} 

The number 40567 appears as part of a course reference or identifier, not an address or geographic location. There is no accompanying city, state, or postal reference to support interpreting it as a ZIP code.

\end{itemize}

\subsection{Lease Contract Review}

This dataset from \citet{leivaditi2020benchmarkleasecontractreview} contains long excerpts from legal and commercial lease agreements. The task is to extract structured contract information from each document, including key entities such as the: parties involved (lessor and lessee), lease terms, payment conditions, property details, and important contractual dates (start and end dates, renewal clauses, and payment schedules). 
We found this dataset to be full of annotation errors and inconsistencies. Below are some examples of issues we found in this benchmark.

\subsubsection{Issue \#1: Inconsistent lease end\_date annotation}

There is inconsistency in how \texttt{end\_date} is annotated in the ground truth. In some cases, the label contains non-date text (e.g., descriptive phrases or clauses), while in others, it fails to normalize or compute the actual end date from clearly stated terms. In addition, when multiple end dates are possible or implied, it is unclear which dates to include in the extraction.
Here are some examples:

\begin{itemize}
\item \emph{``Article 3 – DURATION The first paragraph shall be modified as follows: "The current lease has been granted and agreed upon for the duration of nine consecutive years and two full months, which shall begin from the effective date as stipulated in Chapter V2. It shall end under the conditions described in Article L.145.9 of the Commercial Code (formerly Article 5 of the September 30, 1953 decree) … Effective date for the current lease: JANUARY 1, 2006 …''} \\  
  $\rightarrow$ \texttt{end\_date: [`It shall end under the conditions described in Article L.145.9 of the Commercial Code (formerly Article 5 of the September 30, 1953 decree)']} 

The ground truth extracts a legal clause instead of a date, even though the effective date (January 1, 2006) and duration (nine years and two months) are explicitly given elsewhere, allowing the true end date to be inferred. It is unclear whether the task instructions expect the model to infer such an end date from the duration or to omit it if not explicitly written. Either way, the extracted text in this ground truth annotation is incorrect, since it is neither a valid date nor an empty value.
  
\item \emph{``Article 3 Lease term and the Lessee's purchase option The lease term of this Contract is five years, from January 1, 2009 to December 31, 2014 … Certificate for Registration and Record of Chongqing Housing Lease Contract … The lease term is from January 1, 2009 to December 31, 2013 …''} \\  
  $\rightarrow$ \texttt{end\_date: [`December 31, 2014']} 

The ground truth keeps December 31, 2014, ignoring the December 31, 2013 date from the official registration certificate, without specifying why that source was deprioritized.

\item \emph{``SIXTH: Lease term.- The term for this contract is of five years starting on the 15th of November of 2012 for the Property Section B … and the 15th of October of 2012 for the house …''} \\  
  $\rightarrow$ \texttt{end\_date: [`five years']} 

Here, the ground truth once again extracts a plain text string, and does not infer the actual end dates, or output None.

\end{itemize}

\noindent This inconsistency makes it unclear whether the \texttt{end\_date} field is intended to capture a literal text reference or a calculated contractual end date, and whether the model should abstain when no explicit date is present or when multiple dates appear in the document.

\subsubsection{Issue \#2: Incorrect annotations}

Here are a few examples of mislabeled data:

\begin{itemize}
\item \emph{``on the other, person or corporations whose corporate seat, number of trade registration and trade title are as mentioned in the Annex 1herewith, to be referred to herein after as lessee or tenant … ANNEX-1 SPECIAL PROVISIONS LESSOR Ankara Teknoloji Gelistirme Bölgesi Kurucu ve Isletici Anonim Sirketi LESSEE Trade title Synplicity Arastirma Gelistirme Ltd. Sti.''} \\   
  $\rightarrow$ \texttt{lessee: [`person or corporations whose corporate seat, number of trade registration and trade title are as mentioned in the Annex 1herewith']} 

The ground truth labels the lessee as the entire descriptive clause instead of the actual party name that appears explicitly in Annex-1, which should be ``Synplicity Arastirma Gelistirme Ltd. Sti."
  
\item \emph{``Effective September 1, 2009, the monthly minimum rental (exclusive of any additional charges) due and payable in accordance with the terms of the Lease shall be as follows: September 1, 2009 January 1, 2010 March 1, 2010 January 1, 2011 January 1, 2012 - - - - - December 31, 2009 February 28, 2010 December 31, 2010 December 31, 2011 February 28, 2013 : : : : : \$42,400.00 \$0.00 \$42,400.00 \$43,283.33 \$44,166.67 per month per month per month per month per month …''} \\ 
  $\rightarrow$ \texttt{term\_of\_payment: [`Effective September 1, 2009, the monthly minimum rental (exclusive of any additional charges) due and payable in accordance with the terms of the Lease shall be as follows:', `September 1, 2009', `January 1, 2010', `March 1, 2010', `January 1, 2011', `January 1, 2012']} 

The ground truth incorrectly extracts only a series of dates rather than the complete payment schedule. The original text is difficult to read due to table-like formatting, but the annotation still fails to capture the intended structure. Instead of representing the full payment terms with corresponding date ranges and amounts, it lists only the start dates of each period, losing the actual information regarding the payment term.

\end{itemize}

\clearpage
\section{Additional Results}
\label{app:moreresults}

We additionally assess the same \emph{per-document} and \emph{per-field} trustworthiness scores under different evaluation metrics beyond AUROC.  As before, when evaluating per-document scores, we consider a LLM output incorrect if any of its fields is wrong. When evaluating per-field scores, we consider the correctness of each field in an output independently of the other fields/outputs.

\subsection{Evaluating Trustworthiness Scores via the Precision @ Num-Errors metric}

We first compare trustworthiness scoring methods via an alternative metric to AUROC.
The \emph{Precision @ Num-Errors} metric (i.e. \emph{Precision @ K} where $K$ = the true number of LLM errors in the dataset) evaluates how accurately each scoring method flags errors.  It answers the practical question: \emph{``If you can only review the top-$K$ documents (or fields for the per-field scores) flagged with lowest score, what proportion of these top-$K$ are actually erroneous?''}  
While AUROC considers precision/recall across all possible score-thresholds, Precision @ Num-Errors solely considers precision at one natural score-threshold (where the number of predicted errors matches the true underlying number).

\begin{figure*}[tbh!]
\centering
  \includegraphics[width=1.06\linewidth]{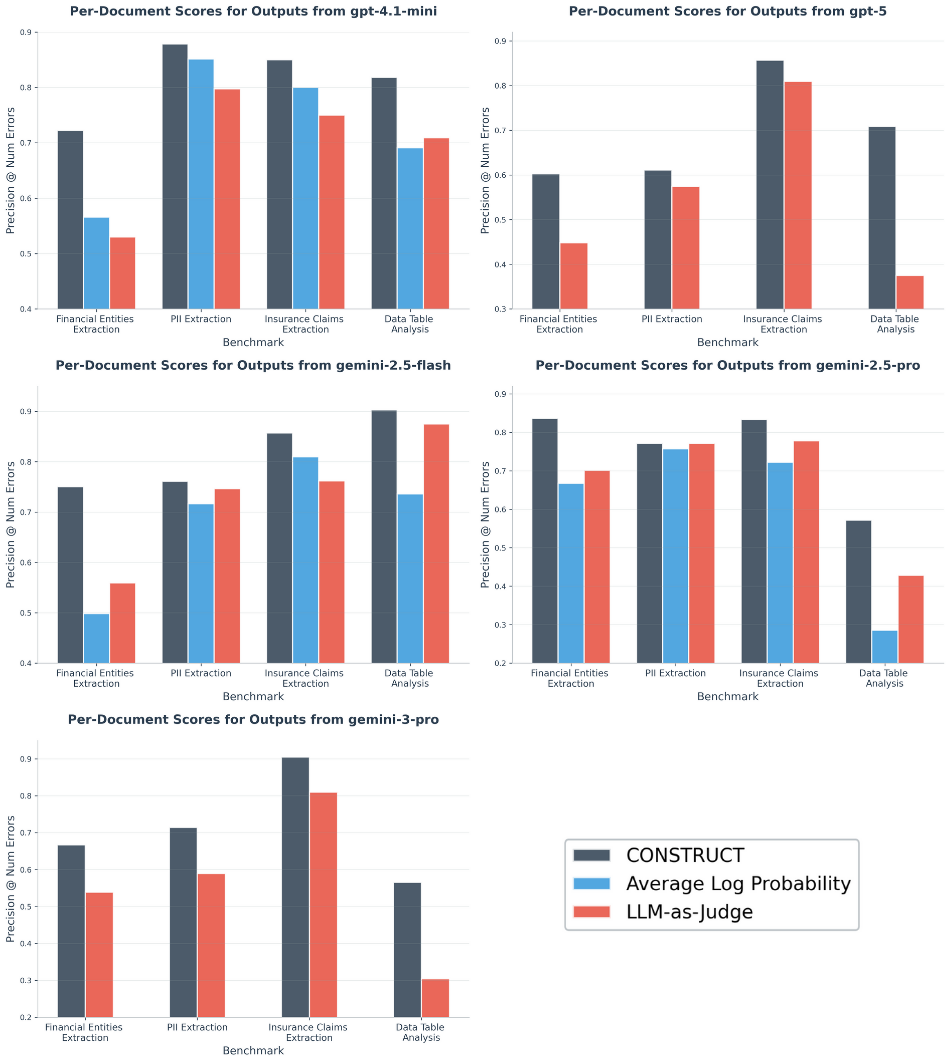}
  \caption{\textbf{Precision @ Num-Errors} achieved by various \textbf{Per-Document} scores applied to outputs generated from different models. Here higher values are better.}
  \label{fig:precision_doc}
\end{figure*}

\begin{figure*}[tbh!]
\centering
  \includegraphics[width=1.06\linewidth]{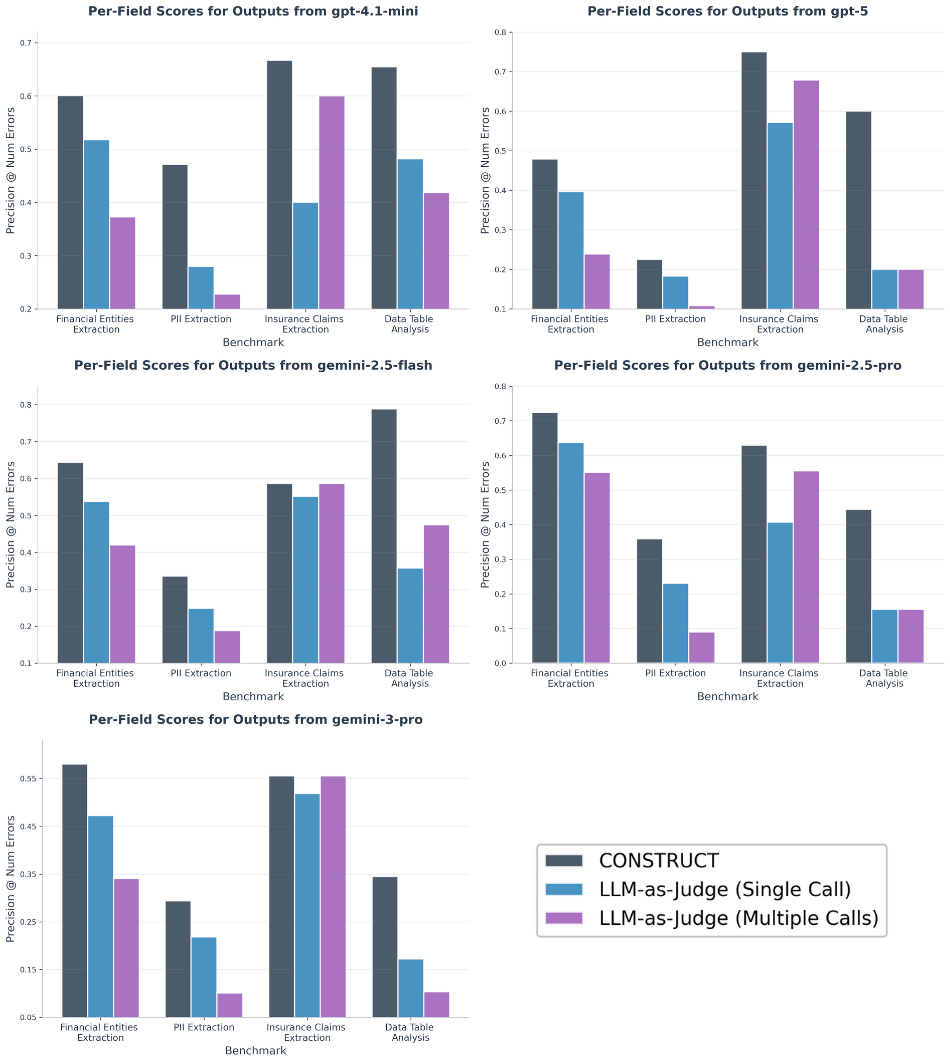}
\caption{\textbf{Precision @ Num-Errors} achieved by various \textbf{Per-Field} scores applied to outputs  generated from different models. Here higher values are better.}
  \label{fig:precision_field}
\end{figure*}

\subsubsection{Per-Document Scoring Results}

Figure \ref{fig:precision_doc} shows that, across all models and benchmarks, CONSTRUCT achieves the highest Precision @ Num-Errors, surfacing more incorrect documents with fewer inspections. Average log probability is somewhat inconsistent: strong for some models/benchmarks but weak for others, but always trails behind CONSTRUCT alongside LLM-as-a-Judge.

\subsubsection{Per-Field Scoring Results}

Figure \ref{fig:precision_field} shows that CONSTRUCT's trust scores consistently achieve the highest Precision @ Num-Errors for field-level error detection across every benchmark and model. The multiple-call LLM-as-a-Judge generally outperforms the single-call version but never matches CONSTRUCT, and does so with much greater computational cost.

\clearpage
\subsection{Evaluating Trustworthiness Scores via the Confidence Gap metric}

We also compare trustworthiness scoring methods under another metric.
The \emph{Confidence Gap} metric measures the margin between scores given to correct vs incorrect outputs -- defined as the difference between the average score of a correct v.s. incorrect output. Unlike Precision/AUROC metrics which are based on relative score rankings across a dataset, this metric also considers the magnitude of scores.  An ideal trustworthiness score should give much lower scores to incorrect outputs, so that humans can interpret raw score magnitudes rather than solely understanding scores on a relative basis.

\begin{figure*}[tbh!]
\centering
  \includegraphics[width=1.06\linewidth]{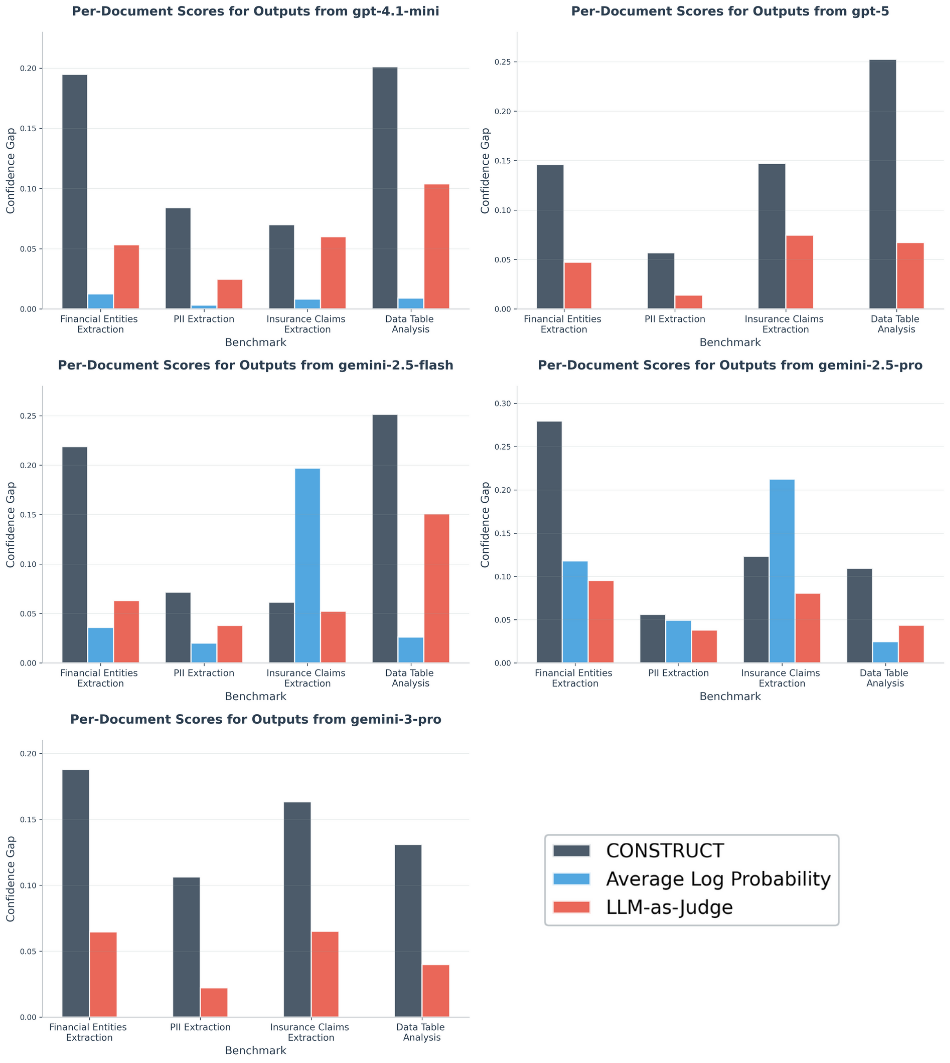}
  \caption{\textbf{Confidence Gap} achieved by various \textbf{Per-Document} scores applied to outputs  generated from different models. Here higher values are better.}
  \label{fig:confidencegap_doc}
\end{figure*}

\begin{figure*}[tbh!]
\centering
  \includegraphics[width=1.06\linewidth]{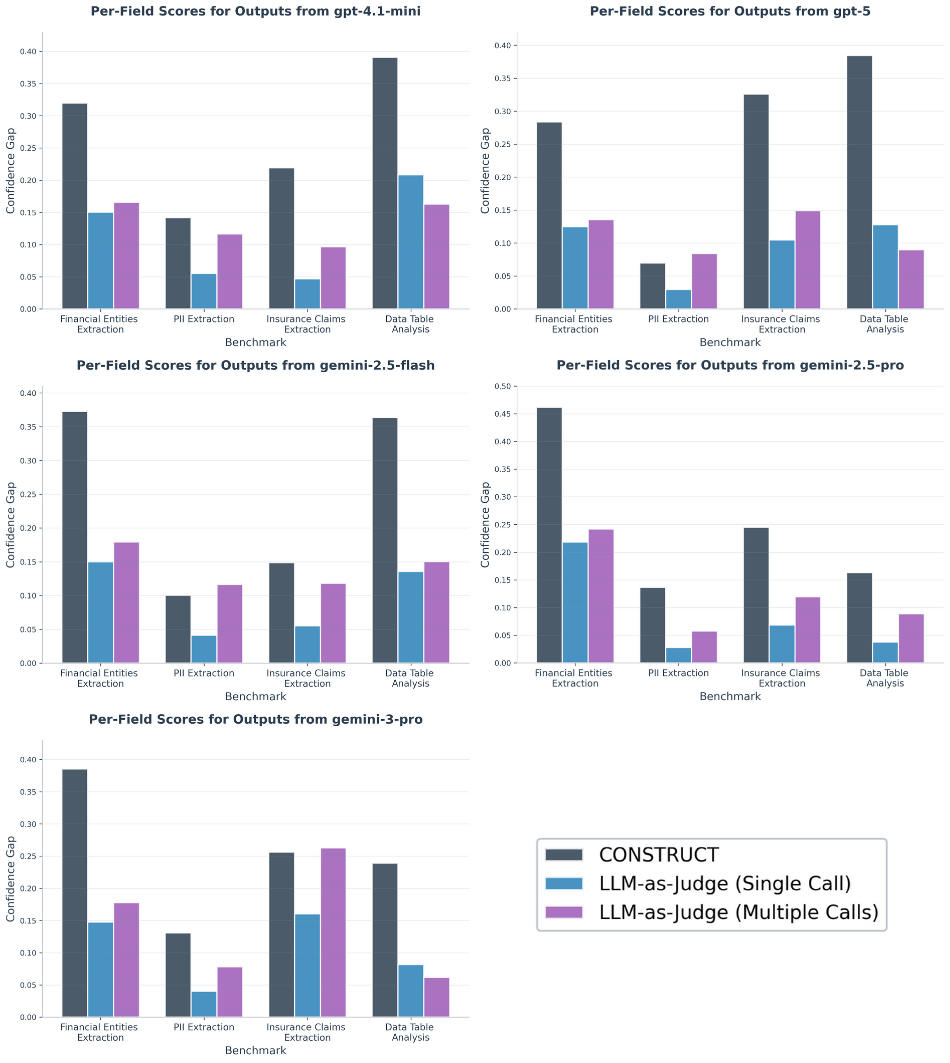}
\caption{\textbf{Confidence Gap} achieved by various \textbf{Per-Field} scores applied to outputs  generated from different models. Here higher values are better.}
  \label{fig:confidencegap_field}
\end{figure*}

\subsubsection{Per-Document Scores}
Results for \emph{per-document} scoring (Figure \ref{fig:confidencegap_doc}) show that:  CONSTRUCT achieves the best Confidence Gap on most models/benchmarks, demonstrating sharper separation between correct and incorrect outputs than other scoring methods. Perplexity-based scoring is highly inconsistent, performing better on the Gemini models but significantly worse on GPT models, whereas LLM-as-a-Judge yields only moderate separation.

\subsubsection{Per-Field Scores}
Results for \emph{per-field} scoring (Figure \ref{fig:confidencegap_field}) show that:   CONSTRUCT produces the largest Confidence Gap across most models and datasets, giving the clearest signal for field-level correctness. The multiple-call LLM-as-a-Judge method improves over single-call, but CONSTRUCT consistently outperforms both (and achieves significantly better efficiency that the multiple-call method).

\clearpage
\section{Additional Details about CONSTRUCT} \label{app:prompt-template}

Below are the \textbf{prompt templates} used for each of CONSTRUCT's five verifier LLM calls outlined in Section \ref{sec:algo}.  
Ablation studies show that individually removing one of the five verifier LLM calls from CONSTRUCT reduces aggregate AUROC error-detection performance over our benchmarks by: 4.4\% for Template \ref{sec:docleveltemplate}, 3.2\% for Template \ref{sec:firstfieldleveltemplate}, 3.6\% for Template \ref{sec:secondfieldleveltemplate}, 3.8\% for Template \ref{sec:firsthybridtemplate}, 5.7\% for Template \ref{sec:secondhybridtemplate} (percentage changes reported on a relative basis).
Adding extra verifier LLM calls to CONSTRUCT only yielded marginal gains $<$1\%, across many types of calls that we investigated.

To produce intermediate scores, outputs from each LLM call with the listed template are naturally mapped to range between 0-1.  Our prompt templates are organized in XML format which helps LLMs better follow them. 
In each template, \textbf{\{...\}} denotes a placeholder variable that is replaced for each example. \textbf{\{input to generator LLM\}} represents the instructions to generate the desired output plus any document data to process (i.e.\ it is the user prompt $x$ to the generator LLM).

\subsection{Prompt Template: Document-Level Scoring}
\label{sec:docleveltemplate}

\begin{tcolorbox}[
  colback=gray!10,
  colframe=gray!40,
  boxrule=0.5pt,
  arc=3mm,
  breakable
]
\begin{BreakVerbatim}
    Below is a User Request and the proposed Answer from an untrustworthy AI assistant:
    
    <request>
    \{PLACEHOLDER: input to generator LLM\}
    
    ## Response Format
    
    The response must follow this JSON schema:
    
    <JSON schema>
    \{PLACEHOLDER: schema specified for generator LLM\}
    </json_schema>
    </request>
    
    <answer>
    \{PLACEHOLDER: structured output from generator LLM\}
    </answer>
    
    
    # Instructions
    
    Think critically to provide a comprehensive veritable argument for why the proposed Answer might be incorrect.
    
    Then consider the argument, consider the Answer again, and determine: How certain are you that the proposed Answer is correct?
    
    Provide a score between 0-100 indicating your confidence that the Answer is correct and factually accurate.
    If the proposed Answer is obviously incorrect, then rate a score of 0.
    
    Your output should strictly use the following template:
    
    
    <think>
    [argue why the proposed Answer might be incorrect (no more than 300 words)]
    </think>
    
    <score>
    [choose a confidence score between 0-100]
    </score>
\end{BreakVerbatim}
\end{tcolorbox}

\subsection{Prompt Template: Field-Level Scoring Template (Numeric Assessment)}
\label{sec:firstfieldleveltemplate}

\begin{tcolorbox}[
  colback=gray!10,
  colframe=gray!40,
  boxrule=0.5pt,
  arc=3mm,
  breakable
]
\begin{BreakVerbatim}
    You are a trustworthy verifier of AI-generated JSON responses.
    Below is a User Request and the proposed JSON Response from an untrustworthy AI assistant.
    Your task is to double check whether each field in the proposed JSON Response is correct.

    
    <request>
    \{PLACEHOLDER: input to generator LLM\}
    
    ## Response Format
    
    The response must follow this JSON schema:
    
    <json_schema>
    \{PLACEHOLDER: schema specified for generator LLM\}
    </json_schema>
    </request>

    <response>
    \{PLACEHOLDER: structured output from generator LLM\}
    </response>
    
    
    ## Instructions
    
    Help me determine how much I can trust the value of each field in the proposed JSON Response.
    It is crucial that you help me catch any inaccurate values.
    
    For each top-level key in the proposed JSON Response:
    - How certain are you that its value in the proposed JSON Response is entirely correct?
    - Provide a score between 0-100 for this key, indicating your confidence that its value is correct and fully accurate.
    - If the value is obviously incorrect, then assign a score of 0 to this key.
    
    Be strict when scoring:
    - Do NOT give high scores for partially correct values, incomplete information, values which are missing but should not be, or content that should not be there.
    - If you cannot verify the accuracy of a value, then your corresponding score should never exceed 50.
    - Only assign high scores if this field in the proposed JSON Response is fully correct and complete.
    - Any issue you identify or verification uncertainty should reduce your score.
    
    ## Output Format
    
    Output a JSON object with the same top-level keys as the proposed JSON Response.
    
    Each top-level key should map to an object with two fields:
    - explanation: argue why the value in the proposed JSON Response might be incorrect
    - score: confidence score between 0-100
\end{BreakVerbatim}
\end{tcolorbox}

\subsection{Prompt Template: Field-Level Scoring Template (Likert Assessment)}
\label{sec:secondfieldleveltemplate}
\begin{tcolorbox}[
  colback=gray!10,
  colframe=gray!40,
  boxrule=0.5pt,
  arc=3mm,
  breakable
]
\begin{BreakVerbatim}
    Below is a User Request and the proposed Response from an untrustworthy AI assistant.
    Your task is to double-check the accuracy of each field in the Response JSON.
    
    <request>
    \{PLACEHOLDER: input to generator LLM\}
    
    ## Response Format
    
    The response must follow this JSON schema:
    
    <json_schema>
    \{PLACEHOLDER: schema specified for generator LLM\}
    </json_schema>
    </request>
    
    <response>
    \{PLACEHOLDER: structured output from generator LLM\}
    </response>
    
    
    ## Instructions
    
    For each top-level key in the JSON:
    Consider whether its value in the proposed JSON Response above seems: potentially incorrect, hard for you to verify based on available information, or missing when it shouldn\'t be.
    Then decide how confident you are that its value is actually correct, choosing one confidence rating out of these choices:
    - Certain
    - Mostly Certain
    - Somewhat Certain
    - Uncertain
    - Likely Incorrect
    
    
    ## Output Format
    
    Output a JSON object with the same top-level keys as the proposed JSON Response.
    
    Each top-level key should map to an object with two fields:
    - explanation: Briefly explain why the value in the proposed JSON Response could be incorrect, incomplete, difficult to verify, or why you are certain it is right
    - confidence: One choice out of: [Certain, Mostly Certain, Somewhat Certain, Uncertain, Likely Incorrect]
\end{BreakVerbatim}
\end{tcolorbox}

LLM Calls using the above template yield one of 5 possible values, which we subsequently map to numeric scores $S_i$ as follows:
Certain $\rightarrow$ 1, Mostly Certain $\rightarrow$ 0.75, Somewhat Certain $\rightarrow$ 0.5, Uncertain $\rightarrow$ 0.25, Likely Incorrect $\rightarrow$ 0.

\subsection{Prompt Template: Document-level scoring + indirect field-level scoring (Accuracy Assessment)}
\label{sec:firsthybridtemplate}

\begin{tcolorbox}[
  colback=gray!10,
  colframe=gray!40,
  boxrule=0.5pt,
  arc=3mm,
  breakable
]
\begin{BreakVerbatim}
    You are an evaluator that identifies factual inaccuracies in AI responses.
    Below is a User Request and the Response provided by an untrustworthy AI Assistant.
    Your task is to find and list any fields in the Response that are factually incorrect, incomplete, or unreliable.

    <request>
    \{PLACEHOLDER: input to generator LLM\}
    
    ## Response Format
    
    The response must follow this JSON schema:
    
    <json_schema>
    \{PLACEHOLDER: schema specified for generator LLM\}
    </json_schema>
    </request>

    <response>
    \{PLACEHOLDER: structured output from generator LLM\}
    </response>
    
    
    ## Instructions
    
    Carefully evaluate the factual accuracy of each top-level field in the JSON Response.
    You must identify any fields that are likely incorrect, incomplete, unverifiable, or misleading.
    
    Be extremely strict in your judgment:
    - Treat any missing or partially correct information as potentially incorrect.
    - Penalize any unsupported assumptions, factual errors, or extraneous content.
    - Incomplete fields should be marked as incorrect. Do not excuse missing data, if a field is null/empty when it should contain information, it is incorrect.
    - Even small inaccuracies should cause a field to be flagged as untrustworthy.
    
    ## Output Format
    
    Output a JSON object with three fields:
    1. "explanation": Briefly describe how you evaluated the response, what issues you found, and why certain fields were marked incorrect.
    2. "incorrect_fields": An array of objects containing ONLY the fields that are potentially incorrect. Each object has:
        - "field_name": The name of the incorrect field
        - "explanation": A brief explanation of why this field is incorrect
    If all fields appear accurate and trustworthy, output an empty array ([]).
    3. "confidence_score": A score between 0 and 100 indicating your confidence in the overall correctness of the response. If you identify any incorrect, incomplete, or unverifiable fields, the score should be very low. Only assign a high score if every field is fully accurate and trustworthy.
    
    Think through your evaluation systematically and provide clear reasoning for your decisions.
\end{BreakVerbatim}

\end{tcolorbox}

\subsection{Prompt Template: Document-level scoring + indirect field-level scoring (Confidence Assessment)}
\label{sec:secondhybridtemplate}

\begin{tcolorbox}[
  colback=gray!10,
  colframe=gray!40,
  boxrule=0.5pt,
  arc=3mm,
  breakable
]
\begin{BreakVerbatim}
    Below is a User Request and the proposed Response from an untrustworthy AI assistant:

    <request>
    \{PLACEHOLDER: input to generator LLM\}
    
    ## Response Format
    
    The response must follow this JSON schema:
    
    <json_schema>
    \{PLACEHOLDER: schema specified for generator LLM\}
    </json_schema>
    </request>

    <response>
    \{PLACEHOLDER: structured output from generator LLM\}
    </response>
    
    
    How can I be sure each field in this response is correct?
    
    Please provide a detailed, field-by-field evaluation of the proposed Response.
    
    
    For each top-level field:
    - Describe the evidence supporting the field’s content.
    - Assess the reliability of that evidence and explain why it can or cannot be trusted.
    - Identify any weaknesses: note if the field appears incorrect, incomplete, misleading, or unverifiable.
    
    Work step by step: systematically check each field for factual accuracy and completeness, then summarize your findings.
    
    This field-level evaluation should directly guide your overall confidence rating - the more issues you find, the lower the rating should be.
    Now rate your confidence on a scale of 0-10 that the response is correct.
    
    
    ## Output Format
    
    Output a JSON object with three fields:
    1. "explanation": Briefly describe how you evaluated the response, what issues you found, and why certain fields were marked incorrect.
    2. "incorrect_fields": An array of objects containing ONLY the fields that are potentially incorrect. Each object has:
        - "field_name": The name of the incorrect field
        - "explanation": A brief explanation of why this field is incorrect
    If all fields appear accurate and trustworthy, output an empty array ([]).
    3. "rating": A rating between 0 and 10. If you identify any incorrect, incomplete, or unverifiable fields, the rating should be very low. Only assign a high rating if every field is fully accurate and trustworthy.
\end{BreakVerbatim}
\end{tcolorbox}

\subsection{Explaining why Outputs seem untrustworthy}
\label{sec:explanations}

To automatically explain \emph{why} a LLM structured output was given a low trustworthiness score, CONSTRUCT first considers which of the verifier LLM Calls yielded the lowest intermediate score. We then return the value of the \texttt{"explanation"} field in the corresponding verifier output (see the aforementioned prompt templates for a definition of this field).  When explaining per-field trustworthiness scores, we only consider the verifier LLM calls that yield intermediate field-level scores $S_i$.

\newpage
\section{Additional Details about our Experiments}
\label{app:experimentdetails}

Here are the \textbf{prompt templates} we used for LLM-as-a-Judge.
They are directly adapted from the LLM-as-a-Judge prompt from  \citet{zheng2023judgingllmasajudgemtbenchchatbot}, which was found to perform nearly as well as human reviewers.
Our LLM-judge to assess a structured output takes in: a user request containing all the relevant information (including document to process), the desired Structured Output schema, and the generator model’s Structured Output response to this request.

\subsection{Per-Document Score LLM-as-a-Judge Prompt Template}

\begin{tcolorbox}[
  colback=gray!10,
  colframe=gray!40,
  boxrule=0.5pt,
  arc=3mm
]
[System] \\
Please act as an impartial judge and evaluate the quality of the response provided by an AI assistant to the user question displayed below. Your evaluation should consider factors such as the helpfulness, relevance, accuracy, depth, creativity, and level of detail of the response. \\
Begin your evaluation by providing a short explanation. Be as objective as possible. After providing your explanation, please rate the response on a scale of 1 to 10 by strictly following this format: ``[[rating]]", for example: ``Rating: [[5]]". \\

[Question] \\
\{PLACEHOLDER: input to generator LLM\} \\

[The Start of Assistant's Answer] \\
\{PLACEHOLDER: output from generator LLM\} 

[The End of Assistant's Answer] 
\end{tcolorbox}

\noindent \textbf{Definition of template variable: \texttt{\{PLACEHOLDER: input to generator LLM\}}} 
\\[-2em]
\begin{tcolorbox}[
  colback=gray!10,
  colframe=gray!40,
  boxrule=0.5pt,
  arc=3mm
]
\{PLACEHOLDER: user prompt to generator LLM, including document to process\} \\

\#\# Response Format \\ 
Your output must be formatted using the following JSON schema: \\

\textless{}json\_schema\textgreater{} \\
\{PLACEHOLDER: structured output schema supplied to   generator (here as string)\} \\
\textless{}/json\_schema\textgreater{}
\end{tcolorbox}
\noindent Throughout, this variable includes the user request, document to process, and generator's desired Structured Output schema -- all concatenated together as  sections of a string.

\subsection{Per-Field Score Single Call LLM-as-a-Judge Prompt Template}

\begin{tcolorbox}[
  colback=gray!10,
  colframe=gray!40,
  boxrule=0.5pt,
  arc=3mm
]
[System] \\
Please act as an impartial judge and evaluate the quality of the response provided by an AI assistant to the user question displayed below. Your evaluation should consider factors such as the helpfulness, relevance, accuracy, depth, creativity, and level of detail of the response. \\
Begin your evaluation by providing a short explanation. Be as objective as possible. After providing your explanation, please rate the response on a scale of 1 to 10. \\

[Question] \\
\{PLACEHOLDER: input to generator LLM\} \\

[The Start of Assistant's Answer] \\
\{PLACEHOLDER: output from generator LLM\} 

[The End of Assistant's Answer] 
\end{tcolorbox}

This LLM-as-a-Judge is required to return a JSON Structured Output object where each field corresponds to a field within the generator's original Structured Output and contains an explanation and a rating of that field.
We provide this LLM-Judge with the following Pydantic schema to enable it to output structured per-field scores.

\begin{lstlisting}[language=Python, style=output] 
class Rating(BaseModel):
    explanation: str
    rating: conint(ge=0, le=10)

def make_pydantic_model(name: str, keys: list[str]) -> type[BaseModel]:
    fields = {key: Rating for key in keys}
    return create_model(name, **fields)
\end{lstlisting}

We create the actual desired JSON schema provided to the LLM-Judge by calling \texttt{make\_pydantic\_model()} with the list of field-names as \texttt{keys}, and then pass supply as an auxiliary input to this structured output LLM-Judge call. 
Below is the expected structured response format from this LLM-as-a-Judge:

\begin{lstlisting}[style=output]
{
    'key_1': {
        'explanation': str
        'rating': int
    },
    'key_2': {
        'explanation': str
        'rating': int
    },
    ...
}
\end{lstlisting}

\subsection{Per-Field Score Multiple Call LLM-as-a-Judge Prompt Template}

\begin{tcolorbox}[
  colback=gray!10,
  colframe=gray!40,
  boxrule=0.5pt,
  arc=3mm
]
[System] \\
Please act as an impartial judge and evaluate the quality of the response provided by an AI assistant to the user question displayed below. Your evaluation should consider factors such as the helpfulness, relevance, accuracy, depth, creativity, and level of detail of the response.  \\
Begin your evaluation by providing a short explanation. Be as objective as possible. After providing your explanation, please rate the response on a scale of 1 to 10 by strictly following this format: ``[[rating]]", for example: ``Rating: [[5]]". \\

Ignore all fields in the response except for the following field: \{PLACEHOLDER: relevant field\} \\
Provide your explanation and rating for this field and nothing else. \\

[Question] \\
\{PLACEHOLDER: input to generator LLM\} \\

[The Start of Assistant's Answer] \\
\{PLACEHOLDER: output from generator LLM\} 

[The End of Assistant's Answer] 
\end{tcolorbox}


\end{document}